\documentclass{article}

\usepackage{microtype}
\usepackage{graphicx}
\usepackage{subfigure}
\usepackage{booktabs} % for professional tables

\usepackage{hyperref}

\usepackage[accepted]{icml2024}
\usepackage{amsmath}
\usepackage{amssymb}
\usepackage{mathtools}
\usepackage{amsthm}

\usepackage[capitalize,noabbrev]{cleveref}

\usepackage{wrapfig}
\usepackage{xspace}
\usepackage{multirow}
\usepackage{diagbox}
\usepackage{pifont}
\usepackage{enumitem}
\usepackage{arydshln}
\usepackage{adjustbox}

\theoremstyle{plain}

\theoremstyle{definition}

\theoremstyle{remark}

\usepackage[textsize=tiny]{todonotes}

\usepackage{xcolor}

\icmltitlerunning{Preference Optimization for Molecule Synthesis with Conditional Residual Energy-based Models}

\begin{document}

\twocolumn[
\icmltitle{Preference Optimization for Molecule Synthesis with Conditional Residual Energy-based Models}



\icmlsetsymbol{equal}{*}

\begin{icmlauthorlist}
\icmlauthor{Songtao Liu}{psu}
\icmlauthor{Hanjun Dai}{deepmind}
\icmlauthor{Yue Zhao}{usc}
\icmlauthor{Peng Liu}{psu}
\end{icmlauthorlist}

\icmlaffiliation{psu}{The Pennsylvania State University}
\icmlaffiliation{deepmind}{Google DeepMind}
\icmlaffiliation{usc}{University of Southern California}

\icmlcorrespondingauthor{Songtao Liu}{skl5761@psu.edu}

\icmlkeywords{Machine Learning, ICML}

\vskip 0.3in
]



\printAffiliationsAndNotice{}  % leave blank if no need to mention equal contribution

\begin{abstract}
Molecule synthesis through machine learning is one of the fundamental problems in drug discovery. Current data-driven strategies employ one-step retrosynthesis models and search algorithms to predict synthetic routes in a top-bottom manner. Despite their effective performance, these strategies face limitations in the molecule synthetic route generation due to a greedy selection of the next molecule set without any lookahead. Furthermore, existing strategies cannot control the generation of synthetic routes based on possible criteria such as material costs, yields, and step count. In this work, we propose a general and principled framework via conditional residual energy-based models (EBMs), that focus on the quality of the entire synthetic route based on the specific criteria. By incorporating an additional energy-based function into our probabilistic model, our proposed algorithm can enhance the quality of the most probable synthetic routes (with higher probabilities) generated by various strategies in a plug-and-play fashion. Extensive experiments demonstrate that our framework can consistently boost performance across various strategies and outperforms previous state-of-the-art top-1 accuracy by a margin of 2.5\%. Code is available at \url{https://github.com/SongtaoLiu0823/CREBM}.

\end{abstract}

\section{Introduction}
\label{sec:intro}
Machine learning for molecule synthesis is crucial for drug discovery, as it ensures that the new molecule can be integrated into industrial-scale chemical manufacturing processes. Existing data-driven strategies employ retrosynthesis prediction, a concept introduced by~\citet{corey1969computer} and further developed in~\citet{corey1991logic}, to derive precursors from a product, which formalizes multi-step retrosynthetic planning. During each planning step, the one-step retrosynthesis model~\citep{dai2019retrosynthesis,chen2020retro,chen2021deep} generates multiple reactant sets. The search algorithm~\citep{segler2018planning,chen2020retro} then chooses promising sets to further expand synthetic routes. This process is repeated until all molecules at the leaf nodes of the synthetic route\footnote{Note that the definition of a synthetic route in chemistry is different from its counterpart in computer science. In organic synthesis, the term ``route'' typically denotes an organic synthesis flowchart in Figure~\ref{fig:route}. However, in graph theory, a route is defined as a continuous sequence of edges in a graph that connects one vertex to another. In this work, we follow the definition of route in chemistry to avoid misunderstanding.} are commercially available starting materials.

Retrosynthetic planning is still a very challenging problem, which requires us to consider the quality of synthetic routes based on specific criteria. According to Chapter 8 in~\citet{hoffmann2009elements}, possible criteria for evaluating synthetic routes include: ``\emph{\textbf{1.} the shortest route (time involved); \textbf{2.} the cheapest route (cost of materials); \textbf{3.} the novelty of the route (patentability); \textbf{4.} the greenest route (avoidance of problematic waste)}''. However, current evaluation metrics~\citep{chen2020retro,maziarz2023re} often neglect these criteria, primarily focusing on assessing a strategy's performance by the proportion of molecules for which the shortest possible synthetic route, with all leaf nodes as starting materials, can be found within a specified number of iterations. Additionally, one of the most serious issues within these metrics is their failure to verify whether the predicted starting materials are actually capable of undergoing the required reactions to synthesize the target molecule~\citep{liu2023fusionretro,tripp2024retro}. To tackle this challenge and ensure that evaluation metrics more accurately evaluate the model's ability to predict feasible synthetic routes, \citet{liu2023fusionretro} has introduced the set-wise exact match of predicted starting materials with those in reference synthetic routes extracted from the reaction database (USPTO-full), while retro-fallback~\citep{tripp2024retro} evaluates the probability that at least one route can be executed in the wet lab. These alternative metrics provide a more realistic reflection of the strategy's performance.

While we recognize the significance of these criteria like the cost of starting materials, the step count, and whether the predicted synthetic routes are feasible, these crucial aspects are currently not incorporated into the predictions made by existing strategies. The generation of synthetic routes is guided by a product of conditional probabilities, each corresponding to a retrosynthesis prediction, where local normalization is applied. These strategies, while general, fail to adequately consider the above criteria and also suffer from the limitations of local normalization. A notable issue in local normalization is the exposure bias, a discrepancy that arises because, at training time, the model is conditioned on the actual synthetic route, whereas at test time, it depends on its own predictions~\citep{ranzato2016sequence,deng2020residual}.

Although advanced search algorithms like beam search (Retro*-0), MCTS~\citep{segler2018planning}, and Retro*~\citep{chen2020retro} provide some improvement through the use of retrosynthesis prediction probability as a prior and by rescoring at the level of the entire synthetic route, the generation process that one step at a time still tends to lack long-range consideration. This shortcoming is mainly due to the reliance on pure probability for route prediction, without considering the cost and feasibility of starting materials in a forward-looking way. A similar issue also arises in recent large language models~\citep{brown2020language,touvron2023llama}. Without alignment~\citep{ouyang2022training,rafailov2023direct}, relying solely on the highest probability for response generation can lead to unsafe outputs. Incorporating alignment helps steer the generated content to conform to certain standards, improving safety. Moreover, some works~\citep{tripp2022re,genheden2022paroutes,liu2023fusionretro,tripp2024retro} report very little performance difference among some different search algorithms. Based on our investigation, we find it non-trivial to extend or retrofit existing strategies to include those criteria.

The above discussion raises an essential question: \emph{Is it possible to improve the quality of synthetic routes generated by various strategies?} We start with an existing probabilistic model $P_{Retro}(\mathcal{T})$, which integrates a one-step retrosynthesis model with a search algorithm as one strategy and wants to improve the quality of synthetic routes. Energy-based models~\citep{hinton2002training,lecun2006tutorial,ranzato2007unified,xie2016theory,grathwohl2019your,sun2021towards} provide \emph{compositionality}~\citep{du2020compositional,du2021unsupervised,du2023reduce}, which means we can add up multiple energy functions to form a new probabilistic model that has the property of each component. Specifically, we want something like
\begin{equation}
    p_{\theta}(\mathcal{T}) \propto \exp \left( \log P_{Retro}(\mathcal{T}) - E_{\theta}(\mathcal{T})  \right).
\end{equation}
By introducing $E_\theta(\mathcal{T})$, we can incorporate additional properties encoded within $E_\theta(\mathcal{T})$, which is defined on the synthetic route, allowing for route evaluation based on various criteria. This allows us to reformulate our approach as a residual EBM~\citep{deng2020residual} and focus on training $E_\theta(\mathcal{T})$ exclusively. Note that our introduced framework directly operates on top of any strategy $P_{Retro}$ without touching its training and our framework differs from~\citet{sun2021towards}, as our energy function is based on specific criteria, whereas theirs is based on reaction prediction probability. Therefore, our model $P_\theta(\mathcal{T})$ can be guided by the energy function $E_\theta(\mathcal{T})$, enabling controllable synthetic route generation based on various criteria.

It is challenging to directly train our EBM via Maximum Likelihood Estimation (MLE) or score matching~\citep{song2021train} due to the difficulty in computing the normalization term. A commonly used framework based on the contrastive divergence~\citep{carreira2005contrastive} requires the samples from the current model, either using gradient-based Markov Chain Monte Carlo methods~\citep{du2019implicit} in continuous spaces or Gibbs sampling and its improved versions~\citep{sun2023discrete} in discrete spaces. However, for the large space of synthetic routes, these methods would be hard to apply due to their slow convergence and difficulty in sampling. Noise Contrastive Estimation (NCE)~\citep{gutmann2010noise,wang2018learning,parshakova2019global} provides an easier way to train the EBM when the ground-truth data is available. However in our setting, we typically don't have the ground-truth synthetic routes as the supervision, but we can possibly get the preference comparisons between different routes. That motivates us to derive a new training paradigm that fits the application in this setting. Inspired by the recent advancements~\citep{ouyang2022training} in large language models (LLMs) that leverage reward models to steer pre-trained language models, we have the potential to derive comparative preferences among different routes tailored to particular criteria. Therefore, we utilize a preference-based loss function for training our model. Given that open-source datasets lack information on yield rates and starting material costs, we use the feasibility of the synthetic routes as our criteria to implement our EBM. Extensive experimental results show our proposed framework can consistently improve the performance of predictions made by a wide range of strategies. Our contribution can be summarized as follows: 
\begin{figure*}[t]
\centering
\includegraphics[width=1.0\textwidth,clip,trim=0 12pt 0 0]{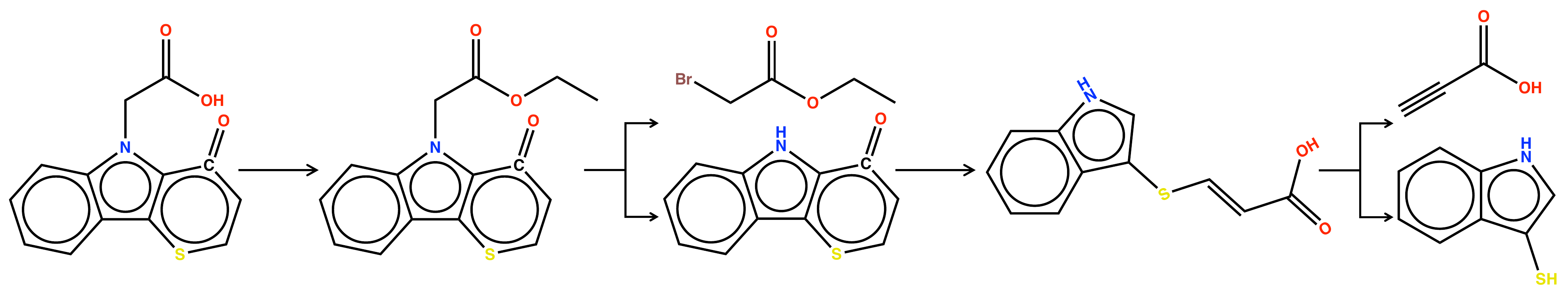}
\vskip -0.05in
\caption{Illustration of a synthetic route. The target molecule we aim to synthesize is the one located on the extreme left, while the molecules positioned at the leaf nodes are the starting materials. The remaining molecules in the diagram are intermediates.
}
\label{fig:route}
\end{figure*}
\begin{itemize}[leftmargin=10pt]
\item We explain retrosynthetic planning through the lens of probability and find that local normalization is present at each planning step. This insight also bridges retrosynthetic planning with sequence generation in LLMs, uncovering certain challenges inherent in retrosynthetic planning (Section~\ref{sec:retrosynthetic planning},~\ref{sec:local normalization}). 

\item Based on this new view, we propose a \textbf{general} and \textbf{principled} framework that can improve the quality of synthetic routes generated by various strategies in a plug-and-play manner (Section~\ref{sec:crebm}).

\item Our work can inspire future research in developing more advanced datasets that consider a broader range of criteria and leverage controllable generation techniques for molecule synthesis (Section ~\ref{sec:crebm}).

\item Extensive experimental results demonstrate the effectiveness of our proposed framework, which can increase previous state-of-the-art top-1 accuracy by 2.5\% (Section~\ref{sec:results}).
\end{itemize}

\section{Preliminary}
In this section, we describe some foundational concepts such as synthetic route and introduce the problem.
\subsection{Synthetic Route}
The chemical molecule space is denoted by $\mathcal{M}$. Starting materials, defined as commercially available molecules, form a subset of $\mathcal{M}$ and are represented by $\mathcal{S} \subset \mathcal{M}$. Note that the term ``building block'' in some literature refers to what we consider starting material. As illustrated in Figure~\ref{fig:route}, a synthetic route comprises a tuple with four elements:
\begin{equation}
    \mathcal{T}=\left(m_{tar}, \tau, \mathcal{I}, \mathcal{B}\right).
\end{equation}
The target molecule is represented by $m_{tar} \in \mathcal{M} \backslash \mathcal{S}$. The starting materials, denoted by $\mathcal{B} \subseteq \mathcal{S}$, undergo a series of reactions $\tau$ to synthesize $m_{tar}$. The intermediates are marked as $\mathcal{I} \subseteq \mathcal{M} \backslash \mathcal{S}$.
\subsection{Reaction \& One-step Retrosynthesis}
We use one injection to denote a one-step reaction as
\begin{equation}
\label{eq:forward}
  \Phi:\mathcal{R} \rightarrow m_p.
\end{equation}
For a product molecule $m_p$, and a set of reactants $\mathcal{R}=\left\{m_r^{(j)}\right\}_{j=1}^n \subseteq \mathcal{M}$ capable of synthesizing $m_p$, with $\Phi(\mathcal{R})=m_p$, we focus on the main product despite potential multiple byproducts in a reaction. The objective of one-step retrosynthesis is to identify a reactant set $\mathcal{R}$ for synthesizing $m_p$. Given that multiple reactant sets may achieve this, indicating a one-to-many relationship, we define $\Psi$ as the mapping from $m_p$ to its various reactant sets $\mathcal{R}_1, \mathcal{R}_2, \ldots, \mathcal{R}_n$, each capable of producing $m_p$.
\begin{equation}
    \Psi: m_p \rightarrow\left\{\mathcal{R}_{1}, \mathcal{R}_{2}, \ldots, \mathcal{R}_{n}\right\}.
\end{equation}
Therefore, $\mathcal{R} \in \left\{\mathcal{R}_{1}, \mathcal{R}_{2}, \ldots, \mathcal{R}_{n}\right\}$ in Eq.~\eqref{eq:forward}. More details about the difference between reaction and retrosynthesis prediction can be found in Appendix~\ref{appendix:diff}. 
\subsection{Retrosynthetic Planning}
The goal of retrosynthetic planning is to find a series of chemical reactions to transform the starting material set $\mathcal{B}\in\left\{\mathcal{B}_{1},\mathcal{B}_{2},\cdots,\mathcal{B}_{n}\right\}$ to a target molecule $m_{tar}\in\mathcal{M}\backslash\mathcal{S}$. Current strategies employ backward chaining, beginning with the target molecule $m_{tar}$ and performing a series of one-step retrosynthesis predictions $\left\{ \Psi_i^{(t)}\right\}_{t=1}^{T_i}\left(\forall i\in \{1, 2,\cdots,n\}\right)$ until all molecules at leaf nodes are from $\mathcal{S}$, which can be formulated as
\begin{equation}
\Gamma: m_{tar} \rightarrow\left\{\mathcal{T}_{1},\mathcal{T}_{2},\cdots,\mathcal{T}_{n}\right\},
\end{equation}
where $\Gamma$ denotes the one-to-many mapping. 
\subsection{Energy-based Models}
Energy-based models define the distribution via an energy function. For $x\in \mathbb{R}^D$, its probability density can be expressed as
\begin{equation}
    P_\theta(x)=\frac{\exp \left(-E_\theta(x)\right)}{Z(\theta)},
\end{equation}
where $E_\theta(x): \mathbb{R}^D \rightarrow \mathbb{R}$ is the energy function, mapping the data point $x$ to a scalar, and $Z(\theta)=\int_{x} \exp \left(-E_\theta(x)\right)$ is the normalization constant. An EBM can be trained using any function that receives data points as input and outputs a scalar value. 

\section{Conditional Residual Energy-based Models for Molecule Synthetic Route Generation}
\label{sec:method}
In this section, we discuss the details of our proposed framework. We first provide a deep understanding of retrosynthetic planning in Section~\ref{sec:retrosynthetic planning} and~\ref{sec:local normalization}, and then discuss the details of our framework in Section~\ref{sec:crebm}.
\subsection{Retrosynthetic Planning is a Conditional Generation Task}
\label{sec:retrosynthetic planning}
Consider a continuous sequence of edges that connects the target molecule to a starting material in a synthetic route. This sequence consists of a series of retrosynthesis predictions, where a reactant in one reaction becomes the product in the subsequent one. This process can thus be conceptualized as a Markov chain, which can be formulated as follows:
\begin{equation}
\label{eq:retro_markov}
\begin{split}
&P_{Retro}\left(m_1, \cdots, m_T \mid m_{tar}\right)\\
=\ &P\left(\Psi\left(m_{tar}\right)\right)\prod_{t=1}^{T-1}  P\left(\Psi\left(m_t\right)\right)\\
=\ &P\left(\mathcal{R}_{tar}\mid m_{tar}\right)\prod_{t=1}^{T-1}  P\left(\mathcal{R}_{t} \mid m_t\right)\\
=\ &P\left(m_1\mid m_{tar}\right)\prod_{t=1}^{T-1}  P\left(m_{t+1} \mid m_t\right),
\end{split}
\end{equation}
where $m_{tar}$ is the target molecule we aim to synthesize, $m_t$ (for $1 \leq t \leq T-1$ ) is the intermediate molecule for which we need to predict its reactant set $\mathcal{R}_{t}$, and $m_T$ is the starting material. In one-step retrosynthesis prediction, all reactants are generated simultaneously during both training and inference, instead of one at a time, so the probability $P\left(\mathcal{R}_{t} \mid m_t\right)$ is equivalent to $P\left(m_{t+1} \mid m_t\right)$, with $m_{t+1} \in\mathcal{R}_{t}$. Although $\mathcal{R}_{tar}$ and $\mathcal{R}_{t}$ $(1\leq t\leq T-2)$ may include starting materials, we exclude these starting materials from the Markov chain for the simplicity of notation since the search ends at these points. In contrast to text generation in an autoregressive manner, where each next token is generated based on all the previous tokens, our one-step retrosynthesis models are trained on individual reactions and only rely on the preceding precursor for the next molecule set generation at each planning step as shown in Eq.~\eqref{eq:retro_markov}. Note that a recent work~\citep{liu2023fusionretro} has conditioned predictions on all predecessor molecules and makes training and inference in an autoregressive manner.
\begin{equation}
P\left(m_{t+1} \mid m_t\right)=P\left(m_{t+1} \mid m_{tar}, m_1, \cdots, m_t\right).
\end{equation}
\subsection{Your Retrosynthesis Model is Secretly a Locally Normalized Model in Retrosynthetic Planning}
\label{sec:local normalization}
The term ``local normalization'' in sequence generation refers to the way probabilities are assigned. To understand this better, consider a language model generating a sequence of tokens. At each step, when the model predicts the next token in the sequence, it considers a set of possible next tokens and assigns probabilities to each of these tokens such that the sum of these probabilities equals one. This process is considered ``local'' because the normalization, which ensures that the probabilities sum to one, is performed independently at each step without considering the entire sequence.

Indeed, in retrosynthetic planning, our retrosynthesis model is locally normalized with a focus on predicting the next reactant set given the current product in the synthetic route. At each planning step, the model evaluates a set of possible reactant sets, normalizing the probabilities of these options so that they sum to one. This normalization happens independently for each retrosynthesis prediction, without considering the entire synthetic route. We explain the details by examining the categorization of retrosynthesis models.
\paragraph{Template-based Model.} In the generation of reactants given a molecule, the widely used template-based model Neuralsym~\citep{segler2017neural} operates by first predicting the template class, which is then applied to the product molecule to derive the reactants. These templates are extracted from the training dataset. Therefore, the retrosynthesis prediction in template-based models can be formulated as follows:
\begin{equation}
   P\left(\mathcal{R}\mid m_p\right)= P\left(Tem|m_p\right),
\end{equation}
where $Tem$ denotes the template. At each step of retrosynthetic planning, Neuralsym assigns probabilities to these pre-defined template options, ensuring they sum to one. This probability reflects how suitable each template is for the current retrosynthesis step, but this assignment is done without taking the entire synthetic route into account. Besides, two other template-based models, GLN~\citep{dai2019retrosynthesis} and RetroComposer~\citep{yan2022retrocomposer}, also apply a softmax function to the scores of all candidate reactants associated with the product. Therefore, some of the widely used template-based models can be considered locally normalized.
\paragraph{Semi-template-based Model.} Semi-template-based (two-stage) models~\citep{shi2020graph,yan2020retroxpert,somnath2021learning} first predict reaction centers and then disconnect the product molecule into synthons which are finally converted into chemically valid molecules. We can consider breaking reaction centers and attaching atoms or leaving groups as actions. Accordingly, we define two action spaces: $\mathcal{E}$ and $\mathcal{A}$. Here, $\mathcal{E}$ denotes the action space for breaking reaction centers and $\mathcal{A}$ represents the action space for attaching atoms or leaving groups. The two-stage model can be formulated as follows:
\begin{equation}
\label{eq:semi-template-based}
\begin{split}
P\left(\mathcal{R} \mid m_p\right) =\ & P\left(\mathcal{G}_{\mathcal{R}} \mid \mathcal{G}_{p}\right)\\
=\ &\sum_{\mathcal{E}, \mathcal{A}} P(\mathcal{E} \mid \mathcal{G}_{p}) P\left(\mathcal{A} \mid \mathcal{G}_{p}, \mathcal{G}_s\right),
\end{split}
\end{equation}
where $\mathcal{G}_{p}=\left(V_{p}, E_{p}\right)$ denotes the product molecule graph with atom set $V_{p}$ and chemical bond set $E_{p}$, $\mathcal{G}_{\mathcal{R}}$ represents the reactants graph, and $\mathcal{G}_s$ denotes the synthons graph. $P\left(\mathcal{R} \mid m_p\right)$ incorporates the combined probabilities of actions in both $\mathcal{E}$ and $\mathcal{A}$ spaces. In the first stage, we predict the reaction centers, based on which we disconnect the product molecule graph. We use a binary label $y_{b} \in\{0,1\}$ for each bond $b\in E_{p}$ in the product molecule graph $\mathcal{G}_{p}$, which indicates whether the bond is a reaction center. So the prediction of the reaction centers can be formulated as
\begin{equation}
   P(\mathcal{E} \mid \mathcal{G}_{p})=\prod_{b\in E_{p}}P(y_{b}|\mathcal{G}_{p}).
\end{equation}
It follows that $\sum \prod_{b \in E_{p}} P\left(y_b \mid \mathcal{G}_{p}\right)=1$. Note that in G2Gs, the center identification is treated as a multi-class classification problem, where $P(\mathcal{E} \mid \mathcal{G}_{p})=1$. In the second stage, we transform synthons into valid molecules. GraphRetro~\citep{somnath2021learning} attaches leaving groups with
\begin{equation}
P\left(\mathcal{A} \mid \mathcal{G}_{p}, \mathcal{G}_s\right)=\prod_{s\in S}P\left(q_{l_s} \mid \mathcal{G}_{p}, \mathcal{G}_s\right),
\end{equation}
where $S$ is the number of connected components (synthons) and the leaving group $q_{l_s}$ is selected from a pre-computed vocabulary from the training dataset. Therefore, we have $\sum \prod_{s\in S}P\left(q_{l_s} \mid \mathcal{G}_{p}, \mathcal{G}_s\right)=1$. G2Gs modifies the second stage by replacing the attachment of a leaving group with a sequential attachment of multiple atoms, which is not difficult to demonstrate that $P\left(\mathcal{A} \mid \mathcal{G}_{p}, \mathcal{G}_s\right)=1$. With $\sum  P\left(\mathcal{A} \mid \mathcal{G}_{p}, \mathcal{G}_s\right)=1$ and $\sum P(\mathcal{E} \mid \mathcal{G}_{p})=1$, it is evident that each stage of the two-stage model operates with local normalization. Based on the conclusion that any model that is locally normalized is also globally normalized~\citep{variani2022global}. We can conclude
\begin{equation}
    \sum P\left(\mathcal{R} \mid m_p\right) =1,
\end{equation}
and these two semi-template-based models are also locally normalized in retrosynthetic planning. 
\begin{figure}[t]
    \begin{center}
\includegraphics[width=0.479\textwidth]{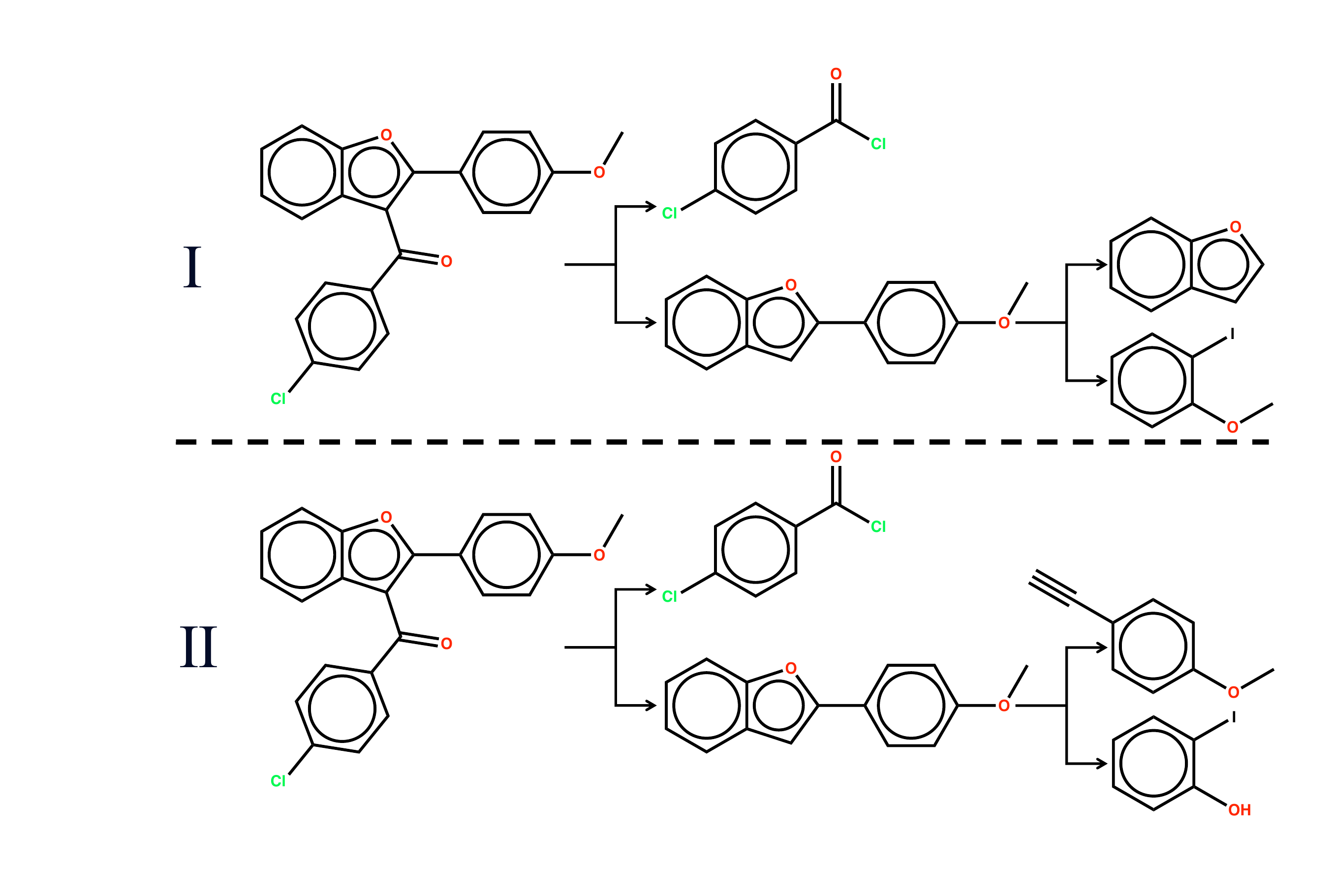}
    \end{center}
    \vskip -0.05in
    \caption{For a given target molecule, we find two synthetic routes that can synthesize it in the dataset.}
   \label{fig:routes}
\end{figure}
\paragraph{Template-free Model.} Template-free models often approach retrosynthesis as a sequence-to-sequence task by representing molecules with SMILES string.
\begin{equation}
    P\left(\mathcal{R} \mid m_p\right)=\prod_{t=1}^T P\left(x_t \mid X_p, x_1, x_2, \cdots,x_{t-1}\right),
\end{equation}
where $x_t$ denotes the $t$-th token in SMILES string of reactants and $X_p$ is the SMILES string of product. Obviously, we have $\sum P\left(x_t \mid X_p, x_1, x_2, \cdots,x_{t-1}\right)=1$ and template-free models are also locally normalized.

For more retrosynthesis models, readers are encouraged to check their implementation and can find that the softmax function is applied to the probability scores of the predicted candidate reactant sets associated with the product. Therefore, the discrete generation nature in retrosynthesis prediction ensures the retrosynthesis model is locally normalized during retrosynthetic planning.

As discussed before, locally normalized models operate on the molecule instead of the whole synthetic route. This step-by-step generation process often fails to account for long-range factors, primarily because it relies on pure probability for predicting routes without forward-thinking various criteria such as starting material costs and the feasibility of synthetic routes. In this work, we propose a general and principled framework via conditional residual energy-based models to improve the quality of synthetic routes. 

\subsection{Conditional Residual Energy-based Models}
\label{sec:crebm}
EBMs offer \emph{compositionality}~\citep{du2020compositional,du2021unsupervised,du2023reduce}, enabling the integration of extra energy functions to evaluate synthetic routes based on multiple criteria and thus develop a new probabilistic model. This formulates our method as a conditional residual EBM~\citep{deng2020residual}.
\begin{equation}
\label{eq:prob_notation}
\begin{split}
    &P_\theta\left(\mathcal{T}\mid m_{tar},c\right)\\
    =\ &P_{Retro}\left(\mathcal{T} \mid m_{tar}\right)\frac{\exp \left(-E_\theta\left(\mathcal{T}\mid m_{tar},c\right)\right)}{Z_\theta\left(m_{tar},c\right)}\\
    \propto \ & P_{Retro}\left(\mathcal{T}\mid m_{tar}\right) \exp \left(-E_\theta\left(\mathcal{T}\mid m_{tar},c\right)\right),
\end{split}
\end{equation}
where $P_{Retro}\left(\mathcal{T}\mid m_{tar}\right)$ is a strategy via the combination of the retrosynthesis model and search algorithm, $c$ denotes specific criteria (condition), $Z_\theta\left(m_{tar}, c\right)$ is a normalizing factor, $P_\theta$ is the joint model, and $E_\theta$ is the conditional residual energy function for evaluating the quality given $c$. During the training of our energy function $E_\theta$, $P_{Retro}\left(\mathcal{T}\mid m_{tar}\right)$ is fixed. Therefore, our approach is a post-training method, freezing the base model when training additional components, and can be applied on top of any existing strategy, while its black-box nature makes it convenient to adopt. With such CREBM, we aim to improve the quality of the most probable synthetic routes, those with higher likelihoods, generated by existing strategies to meet specific criteria and achieve controllable generation.
\subsubsection{Training} 
As discussed in Section~\ref{sec:intro}, maximizing Eq.~\eqref{eq:prob_notation} through MLE is challenging due to the normalization factor. \citet{deng2020residual} address this by sampling sequences from the training dataset and generated sequences from an autoregressive LM as positive and negative pairs, respectively. They then train the energy function as a binary classifier to distinguish between ground truth and generated texts.

However, in retrosynthetic planning, we usually do not have fixed ground truth synthetic routes. Given a product, we can find multiple routes within the dataset to synthesize it as shown in Figure~\ref{fig:routes}. Preference for different synthetic routes varies according to different criteria. For example, consider two synthetic routes: one with a high yield but environmentally unfriendly, and the other with a low yield but eco-friendly. In this case, our preference for either route varies based on our priorities regarding energy conservation and emission reduction. We use a real-world example in Appendix~\ref{appendix:real example} to illustrate the trade-off between yield and environmental impact in molecule synthesis. Therefore, we can't train our model via NCE. Inspired by the recent successes of alignment algorithms~\citep{ouyang2022training,rafailov2023direct} in LLMs, we can potentially obtain preference comparisons among various routes based on specific criteria. Then, we train our model using a preference-based loss function. Given a target molecule $m_{tar}$, we have $n$ synthetic routes $\{\mathcal{T}_1, \mathcal{T}_2, \ldots, \mathcal{T}_n\}$ that can synthesize it. We define a reward function based on a specific criteria $c$.  
\begin{equation}
\label{eq:evaluation func}
    \varphi\left(\cdot\mid m_{tar},c\right): \mathcal{X}_{tar} \rightarrow \mathcal{V},
\end{equation}
where $\mathcal{X}_{tar}$ is the space of synthetic routes for molecule $m_{tar}$ and $\mathcal{V}\in \mathbb{R}$ is a scalar value. The higher the value $\mathcal{V}$ of a synthetic route $\mathcal{T}$, the more it meets our criteria $c$. So given two synthetic routes $\mathcal{T}_1$ and $\mathcal{T}_2$, we can have preference. Bradley-Terry (BT)~\citep{bradley1952rank} model is a common choice for modeling preferences. The likelihood of preferring route $\mathcal{T}_1$ over $\mathcal{T}_2$ is as follows:
\begin{equation}
\begin{split}
    &P^*\left(\mathcal{T}_1 \succ \mathcal{T}_2 \mid m_{tar},c\right)\\
    =\ &\frac{\exp \left(\psi^*\left(\mathcal{T}_1\mid m_{tar},c\right)\right)}{\exp \left(\psi^*\left(\mathcal{T}_1\mid m_{tar},c\right)\right)+\exp \left(\psi^*\left(\mathcal{T}_2\mid m_{tar},c\right)\right)}\\
    &\operatorname{s.t.}\ \varphi\left(\mathcal{T}_1 \mid m_{tar},c\right)>\varphi\left(\mathcal{T}_2 \mid m_{tar},c\right),
    \end{split}
\end{equation}
where $\psi^*$ is the ground truth reward function. $\mathcal{T}_1 \succ \mathcal{T}_2$ means $\varphi\left(\mathcal{T}_1 \mid m_{tar},c\right)>\varphi\left(\mathcal{T}_2 \mid m_{tar},c\right)$, and $\mathcal{T}_1$ is preferred than $\mathcal{T}_2$. 
Thus, the loss function for our model is
\begin{equation}
\label{eq:loss}
\begin{split}
\mathcal{L}&=-\mathbb{E}\left[\log \sigma\left(-E_\theta\left(\mathcal{T}_w\mid m_{tar},c\right)+E_\theta\left(\mathcal{T}_l\mid m_{tar},c\right)\right)\right]\\
&\operatorname{s.t.}\ \left(m_{tar}, \mathcal{T}_w, \mathcal{T}_l\right) \sim \mathcal{D},
\end{split}
\end{equation}
where $\sigma$ is the sigmoid function and $\varphi\left(\mathcal{T}_w \mid m_{tar},c\right)>\varphi\left(\mathcal{T}_l \mid m_{tar},c\right)$ given $c$. We use a reference synthetic route and multiple routes sampled from $P_{Retro}$ for each molecule in the training dataset to construct our synthetic preference dataset $\mathcal{D}$. In our implementation, we only use one strategy (Neuralsym+Retro*-0) as our $P_{Retro}$ for sampling multiple synthetic routes to update $\theta$ during training.
\subsubsection{Implementation of $\varphi$}
\label{sec:implementation of reward function}
The key is the implementation of $\varphi$, which ranks synthetic routes that align with diverse criteria. Some criteria, such as the novelty and reliability of a synthetic route, are not easily quantifiable through a reward function. While some criteria like material costs can be quantified, a pharmaceutical company's production equipment may not necessarily be compatible with the cheapest route predicted by the machine learning model. Therefore, it is crucial to consider whether the available equipment can accommodate the reaction conditions in the predicted synthetic routes. However, incorporating production equipment specifications into the reward function design poses a significant challenge. Ideally, chemists or manufacturers would evaluate and rank the routes according to their expertise, but it is expensive for our research. Therefore, we design a heuristic reward function for ranking purposes, rather than estimating rewards. This approach aligns with recent developments in LLMs, which also rank responses instead of assigning accurate scores when constructing preference datasets.

In this work, we focus on assessing the feasibility of starting materials for synthesizing the target molecule. This emphasis comes from the necessity for machine learning models to first ensure that the predicted synthetic route can be executed in the wet lab before considering other criteria. We use a forward model to simulate the synthesis process to replace the wet lab experiments and compare the similarity of the produced product with the target molecule. Besides, we have multiple routes for synthesizing the same target molecule. Sometimes a new synthetic route can be obtained by simply substituting a chlorine atom in a material with a different halogen atom. Therefore, we also consider routes whose materials are similar to those of the reference route as a better choice. So our $\varphi$ can be rewritten as
\begin{equation}
\label{eq:sim}
   \varphi\left(\mathcal{T}\mid m_{tar},c\right)= \operatorname{sim}\left(f(\mathcal{B}), m_{tar}\right) + \operatorname{sim}\left(\mathcal{B}, \mathcal{B}_{ref}\right),
\end{equation}
where $\mathcal{B}_{ref}$ is the starting material set of the reference synthetic route, $f$ denotes the forward model, and $\operatorname{sim}\left(\cdot, \cdot\right)$ is the Tanimoto similarity function. Note that the Tanimoto similarity function provides a measure of similarity between two sets of fingerprint bits. Therefore, it can be used for computing the similarity between two sets of molecules.

\begin{algorithm}[t]
\caption{CREBM Framework}
\label{alg:inference}
\begin{algorithmic}[1]
\STATE \textbf{[Train Phase]: Learning}:
\STATE Define reward function $\varphi\left(\cdot\mid m_{tar},c\right)$ as Eq.~\eqref{eq:evaluation func}
\STATE $\varphi\left(\cdot\mid m_{tar},c\right): \mathcal{X}_{tar} \rightarrow \mathcal{V}$
\STATE Rank synthetic routes $\mathcal{X}_{tar}\sim P_{Retro}$ based on $\varphi$
\STATE \textbf{Train Conditional Residual Energy-based Models}:
\STATE $\theta^*=\arg\min_\theta\mathcal{L}=\arg \max_\theta \mathbb{E}_{\left(m_{tar}, \mathcal{T}_w, \mathcal{T}_l\right) \sim \mathcal{D}}$
\STATE \textbf{[Test Phase]: Inference}: 
\STATE \textbf{Input}: $\theta^*$, $m_{tar}$, Proposal $\mathcal{X}_{tar} \sim P_{Retro}\left(\cdot\mid m_{tar}\right)$. 
\STATE $L \gets -\log P_{Retro}\left(\mathcal{T}\mid m_{tar}\right) +E_\theta\left(\mathcal{T}\mid m_{tar},c\right)$
\STATE $\mathcal{T}^* = \arg\min_{\mathcal{T} \in \mathcal{X}_{tar}}L$
\STATE \textbf{Return} $\mathcal{T}^*$
\end{algorithmic}
\end{algorithm}
\subsubsection{Inference}
After completing the training, our energy function can be applied on top of any strategy in a plug-and-play manner. With the well-learned energy function $E_{\theta^*}$ where $\theta^*=\arg \min _{\theta} \mathcal{L}$, the inference process aims to find the best synthetic route $\mathcal{T}^*$ that minimizes the negative log-likelihood for a specific target molecule $m_{tar}$ and a given criteria $c$,  i.e.
\begin{equation}
\small{\mathcal{T}^{*}=\arg \min_{\mathcal{T} \in \mathcal{X}_{tar}}\left(E_\theta\left(\mathcal{T}\mid m_{tar},c\right)
-\log P_{Retro}\left(\mathcal{T}\mid m_{tar}\right)\right)}.
\end{equation}
In other words, we find $\mathcal{T}^*$ by maximizing $P_{\theta^*}\left(\mathcal{T}\mid m_{tar},c\right)$ among all routes sampled from $P_{Retro}\left(\mathcal{T}\mid m_{tar}\right)$. We illustrate the overall process in Algorithm~\ref{alg:inference}.
\subsubsection{Discussion}
In this section, we answer some questions that are related to our work.

\textbf{Q1:} With the reward function, why use Bradley-Terry model instead of reinforcement learning for CREBM training? \\
\textbf{A1:} As discussed in Section~\ref{sec:implementation of reward function}, it's very difficult to build an accurate point-wise reward function~\citep{bhattacharyya2020energy,haroutunian2023reranking} for reinforcement learning (RL). We attempted to use the reward function in Eq.~\eqref{eq:evaluation func} to train an RL model. We generated synthetic routes from the RL model and used the product of its negative log-likelihood and the reward function as the training loss. However, experimental results indicated a significant drop in accuracy compared to the base model. Training and fine-tuning an RL model proved to be very challenging. In contrast, we did not adjust any hyperparameters when training our CREBM. Therefore, we believe training with a preference-based loss function is better, or at least much easier to make things work.

\textbf{Q2:} Why not use the parameter-efficient fine-tuning for preference alignment?\\
\textbf{A2:} Directly modifying the base model can be challenging because: \textbf{1.} Route generation involves a single-step retrosynthesis model and a search algorithm, both of which are currently trained independently. Integrating these components for joint training is necessary for parameter-efficient fine-tuning, but this integration requires significant modifications to both parts, which we believe is inefficient. For instance, it is difficult to fine-tune a template-based classifier for preference alignment. \textbf{2.} While specific criteria necessitate fine-tuning all strategies, with CREBM, we only need to train the model for these criteria once. Afterward, it can be applied to any strategy in a plug-and-play fashion. As discussed in Section~\ref{sec:crebm}, we can conclude our CREBM approach is more efficient.

\textbf{Q3:} Is a fair comparison since the baseline methods don't use an additional reward function?\\
\textbf{A3:} Our method is complementary to most existing strategies, as we are building an adapter that can improve most existing methods by treating them as a base model in the CREBM framework. Experimental results in Section~\ref{sec:results} demonstrate that our method consistently enhances the performance across all base methods and is a general model-agnostic approach that can bring additional benefits with this lightweight post-training.

\textbf{Q4:} Comparison with an explicit search policy function like PDVN~\citep{liu2023retrosynthetic}?\\
\textbf{A4:} PDVN requires training different explicit search policy functions based on the specific retrosynthesis models employed. Our approach begins by sampling routes from Neuralsym and Retro*-0. These sampled routes are then utilized to train our CREBM. Subsequently, we employ our CREBM as an adapter to collaborate with any base model and search algorithm during inference, without the need for any additional tuning.

\section{Experiments}
In this section, we evaluate the performance of our proposed framework in retrosynthetic planning.
\begin{table*}[htbp]
\centering
\caption{Summary of retrosynthetic planning results in terms of exact match accuracy (\%). Our framework (CREBM) can consistently improve the performance of existing strategies.}
\label{tab:main results}
\vskip 0.1in
\setlength{\tabcolsep}{0.75mm}
\begin{tabular}{lccccccccccc}
\toprule
\multirow{2}{*}{\diagbox{One-step Model}{Search Algorithm}}  & \multicolumn{5}{c}{Retro*} & \multicolumn{5}{c}{Retro*-0} & \multicolumn{1}{c}{Greedy DFS} \\
\cmidrule(r){2-6} \cmidrule(r){7-11} \cmidrule(r){12-12}
&  Top-1     &  Top-2  &   Top-3  &   Top-4 &   Top-5
&  Top-1     &  Top-2  &   Top-3  &   Top-4 &   Top-5
&   Top-1  \\
\midrule
\multicolumn{1}{c}{Template-based} \\
\midrule
Retrosim~\citep{coley2017computer}  &35.1  &40.5  &42.9   &44.0  &44.6  &35.0  &40.5  &43.0  &44.1 & 44.6 & 31.5 \\
\hdashline
Neuralsym~\citep{segler2017neural}  & 41.7  & 49.2  &52.1   &53.6  &54.4  &42.0  &49.3  &52.0  &53.6 & 54.3 & \textbf{39.2} \\
Neuralsym+CREBM &\textbf{44.2}  &50.8  &53.6  &54.6 & 55.4 & \textbf{44.5}  & \textbf{51.0}  &53.5   &54.5  &55.2  & - \\
\hdashline
GLN~\citep{dai2019retrosynthesis}  &39.6  &48.9  &52.7   &54.6  &55.7  &39.5  &48.7  &52.6  &54.5 &55.6 & 38.0 \\
GLN+CREBM  &43.3  &\textbf{51.1}  &\textbf{53.9}  &\textbf{55.5} & \textbf{56.4} &43.2  &\textbf{51.0}  &\textbf{53.8}   &\textbf{55.5}  &\textbf{56.3}  & - \\
\midrule
\multicolumn{1}{c}{Semi-template-based} \\
\midrule
G2Gs~\citep{shi2020graph}  &5.4  &8.3  &9.9   &10.9  &11.7  &4.2  &6.5  &7.6  &8.3 &8.9  & 3.8 \\
\hdashline
GraphRetro~\citep{somnath2021learning}  &15.3  &19.5  &21.0   &21.9  &22.4  &15.3  &19.5  &21.0 &21.9 &22.2  & 14.4 \\
GraphRetro+CREBM &\textbf{16.3}  &\textbf{20.1}  &\textbf{21.6} &\textbf{22.3} &\textbf{22.7}  &\textbf{16.3}  &\textbf{20.2}  &\textbf{21.6}   &\textbf{22.3}  &\textbf{22.7}   & - \\
\midrule
\multicolumn{1}{c}{Template-free} \\
\midrule
Transformer~\citep{karpov2019transformer}  &31.3  &40.4  &44.7   &47.2  &48.9  &31.2  &40.5  &45.1  &47.3 &48.7  &26.7  \\
Transformer+CREBM  &35.0  &43.4  &46.7  &48.5 &49.7 &34.9  &43.5  &46.6   &48.4  &49.6   &- \\
\hdashline
Megan~\citep{sacha2021molecule}  &18.8  &29.7  &37.2   &42.6  &45.9  &19.5  &28.0  &33.2  &36.4 &38.5  &32.9  \\
\hdashline
FusionRetro~\citep{liu2023fusionretro}  & 37.5 & 45.0 & 48.2  & 50.0 & 50.9 & 37.5  &45.0  &48.3  &50.2 &51.2  &\textbf{33.8}  \\
FusionRetro+CREBM & \textbf{39.4}  &\textbf{46.6}  &\textbf{49.3}  &\textbf{50.7} &\textbf{51.5} & \textbf{39.6} & \textbf{46.7} & \textbf{49.5}  & \textbf{51.0} & \textbf{51.7}   &-  \\
\bottomrule
\end{tabular}
\end{table*}
\subsection{Experimental Setup}
\paragraph{Dataset.} We use the public dataset RetroBench~\citep{liu2023fusionretro} for evaluation. The target molecules associated with synthetic routes are split into training, validation, and test datasets in an 80\%/10\%/10\% ratio. We have 46,458 data points for training, 5,803 for validation, and 5,838 for testing. Synthetic routes for each target molecule are extracted from the reaction network which is constructed with all reactions in USPTO-full. More details about the dataset can be found in Appendix~\ref{appendix:dataset}.
\paragraph{Evaluation Protocol.}
We use the set-wise exact match between the predicted starting material set and that of the reference route in the test dataset as our evaluation metric. Note that for a given target, there might be several synthetic routes available in the test set. We consider the prediction to be accurate when the set of starting materials predicted matches with any one of the various reference options.
\paragraph{Baselines.} A retrosynthetic planning strategy is a combination of a retrosynthesis model and a search algorithm. We consider template-based models: Retrosim~\citep{coley2017computer}, Neuralsym~\citep{segler2017neural} and GLN~\citep{dai2019retrosynthesis}; template-free models: Transformer~\citep{karpov2019transformer} and Megan~\citep{sacha2021molecule}; semi-template-based models: G2Gs~\citep{shi2020graph} and GraphRetro~\citep{somnath2021learning}; as our retrosynthesis models. For the algorithm, we use Retro*~\citep{chen2020retro} and Retro*-0. Retro* utilizes a neural architecture to evaluate the future score in retrosynthetic planning, while Retro*-0 is indeed a beam search.
\paragraph{Implementation Details.} In this work, our focus is on the feasibility of the starting materials. Therefore, we disregard intermediate molecules for our $E_{\theta}$. In a more general case, the preferences between routes can depend on intermediate molecules as well. We leave this exploration to future work, which is flexible in our framework. We employ a standard Transformer~\citep{vaswani2017attention} architecture to implement $E_\theta\left(\mathcal{T}\mid m_{tar},c\right)$, with the target molecule serving as the input for the encoder and the starting material (right shift) as the input for the decoder. The output is the logits of the starting material (left shift) for computing $E_\theta$. One thing we’d like to point out is that $E_\theta$ is pretrained first on the target-to-starting material task, so we naturally deploy this for modeling, instead of training an encoder-only one from scratch. We also employ the standard Transformer architecture to implement the forward model, framing the task of predicting a product from starting materials as a sequence-to-sequence task. For constructing our preference dataset $\mathcal{D}$, we sample 10 synthetic routes for each molecule in the training dataset. All the models in the work are trained on the NVIDIA Tesla A100 GPU. More details about hyper-parameters can be found in Appendix~\ref{appendix:repro}. 
\begin{table}[t]
\centering
\caption{Summary of results with our CREBM in terms of top-1 accuracy (\%) on routes of different depths.}
\label{tab:depth}
\vskip 0.1in
\setlength{\tabcolsep}{2.4mm}
\begin{tabular}{lccccc}
\toprule
\multirow{2}{*}{{Model}}& \multicolumn{5}{c}{Retro*-0}  \\
\cmidrule(r){2-6} 
&  2    &  3  &   4  &   5 &   6 \\
\midrule
Neuralsym  &46.1  &42.0  &33.7  &37.3 & 40.8 \\
+CREBM  & +2.2  & +2.1  &+4.9   &+2.9  &+5.0   \\
\midrule
GLN  &46.4  &39.0  &32.6  &25.3 &21.8  \\
+CREBM  &+2.7  &+3.1  &+4.7   &+7.1  &+14.5   \\
\midrule
Transformer &39.3  &30.1  &20.9  &15.2 &16.2    \\
+CREBM  &+2.4  &+5.3  &+3.0   &+5.6  &+10.6    \\
\bottomrule
\end{tabular}
\end{table}
\subsection{Results}
\label{sec:results}
We report top-k accuracy in Table~\ref{tab:main results}. We reuse the metrics of the baselines already reported in~\citet{liu2023fusionretro}. The results demonstrate that the strategies equipped with our framework, CREBM, achieve better performance. Specifically, our framework can achieve state-of-the-art top-1 accuracy with Neuralsym+Retro*-0. We can conclude our framework can consistently improve the performance of all strategies, and predict more feasible synthetic routes. As shown in Table~\ref{tab:depth}, our framework can also improve performance no matter how long the synthetic route is. When the length of routes is larger, our amplification becomes more pronounced, which demonstrates our framework can improve the performance of long route prediction. 
\subsection{Ablation Study}
We conduct an ablation study with Retro*-0+Neuralsym to verify the effectiveness of our proposed energy function. As shown in Table~\ref{tab:ablation}, only using the energy function for route ranking results in a significant drop in top-1 accuracy and a marginal decrease in top-5 accuracy. This indicates that our energy function still performs well within the top-5 predictions. Adjusting the probability by adding or subtracting the energy function leads to corresponding increases or decreases in accuracy, thereby confirming the effectiveness of our energy function from both positive and negative perspectives. 
\begin{table}[t]
\caption{Ablation study on our energy function.}
\vskip 0.1in
\label{tab:ablation}
\centering
\resizebox{\linewidth}{!}{
\setlength{\tabcolsep}{2pt}
\begin{tabular}{l|c|c|c|c}
\toprule
    Metric for Ranking & Top-1  &  $\Delta_1$ & Top-5 &  $\Delta_2$\\
    \midrule
    $-\log P_{Retro}\left(\mathcal{T}\mid m_{tar}\right)$ & 42.0 & 0 & 54.3 & 0  \\
    \midrule
    $-\log P_{Retro}\left(\mathcal{T}\mid m_{tar}\right) +E_\theta\left(\mathcal{T}\mid m_{tar},c\right)$ & 44.5 & +2.5 & 55.2 & +0.9 \\
$E_\theta\left(\mathcal{T}\mid m_{tar},c\right)$ & 34.1 & -7.9 & 53.1 &-1.2 \\
    \footnotesize{$-\log P_{Retro}\left(\mathcal{T}\mid m_{tar}\right) -E_\theta\left(\mathcal{T}\mid m_{tar},c\right)$} & 19.2 & -22.8 &44.6 &-9.7 \\
    \bottomrule
    \end{tabular}
}
\end{table}
\begin{figure}[t]
    \begin{center}
        \includegraphics[width=0.479\textwidth]{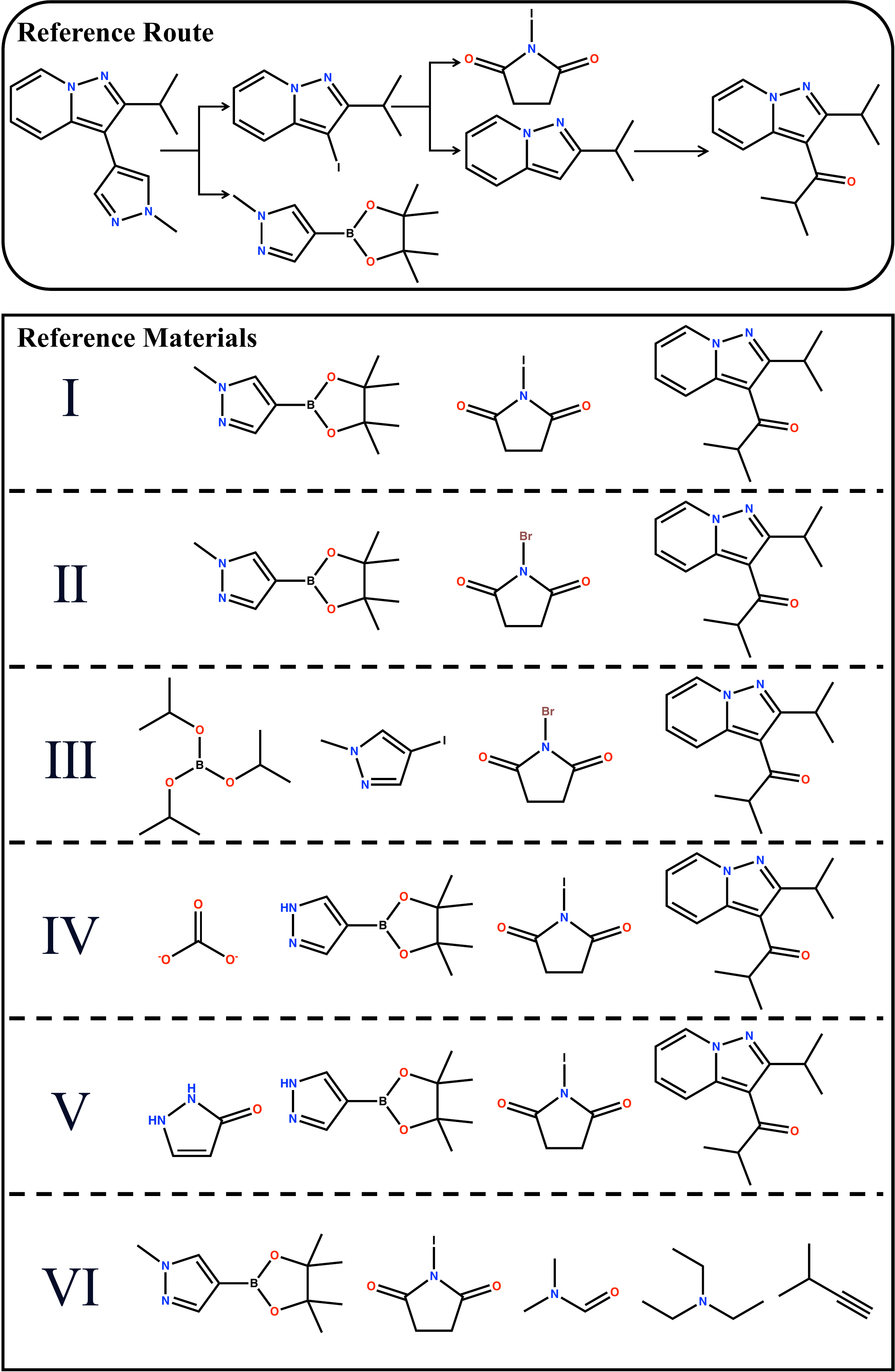}
    \end{center}
    \vskip -0.05in
    \caption{A visual example of a target molecule's synthetic routes used to train the energy function, with these routes ranked according to Eq.~\eqref{eq:sim}.}
   \label{fig:visual}
\end{figure}
\subsection{Visualization of Synthetic Routes used to Train the Energy Function}
We provide a visual example (Figure~\ref{fig:visual}) of the synthetic routes that are used to train the energy function and ranked based on Eq.~\eqref{eq:sim}. We can find that higher-ranked routes have a greater likelihood of successfully synthesizing the target molecule, and their starting materials are more closely aligned with the reference materials. The starting materials of the second-ranked route only require replacing the iodine atom in one of the starting materials of the reference route with a bromine atom.

\section{Related Work}
\paragraph{Retrosynthesis Model.} Current one-step retrosynthesis models fall into three distinct categories: template-based~\citep{coley2017computer,segler2017neural,dai2019retrosynthesis,chen2020retro,chen2021deep,seidl2021modern}, semi-template-based~\citep{shi2020graph,yan2020retroxpert,somnath2021learning}, and template-free~\citep{liu2017retrosynthetic,zheng2019predicting,chen2019learning,karpov2019transformer,ishiguro2020data}.
\paragraph{Search Algorithm.}Search algorithms~\citep{segler2018planning,han2022gnn} select the most promising candidate at each step of the retrosynthetic planning. More details about related work can be found in Appendix~\ref{appendix:related}.

\section{Conclusion and Future Work}
In this work, we propose a framework based on conditional residual energy-based models to improve the quality of synthetic routes generated by existing retrosynthetic planning strategies. We implement one CREBM to enhance the feasibility of routes for synthesizing the target molecule. Extensive experimental results show our proposed framework can improve the accuracy of existing strategies. Our work can inspire future research to develop compositional CREBMs based on multiple criteria for molecule synthesis.

\section*{Acknowledgements}
We thank all the anonymous reviewers for their helpful comments and suggestions. Songtao Liu thanks Zhengkai Tu for his helpful discussions and comments. Songtao Liu also acknowledges the research assistantship provided by Prof. Dongwon Lee during the summer of 2023. Although the project on geometric deep learning, partially conducted during that summer, resulted in failure, Songtao Liu believes Prof. Dongwon Lee's support helps him to continue his Ph.D. studies. Songtao Liu is deeply grateful to Prof. Dongwon Lee, the director of Penn State's IST doctoral programs, for his kindness which makes international students at IST feel at home. Therefore, this work is partially supported by the National Science Foundation under Grant No. 1820609.

\section*{Impact Statement}

This paper presents work whose goal is to advance the field of 
Machine Learning. There are many potential societal consequences of our work, none which we feel must be specifically highlighted here.


\bibliography{example_paper}

\begin{thebibliography}{110}
\providecommand{\natexlab}[1]{#1}
\providecommand{\url}[1]{\texttt{#1}}
\expandafter\ifx\csname urlstyle\endcsname\relax
  \providecommand{\doi}[1]{doi: #1}\else
  \providecommand{\doi}{doi: \begingroup \urlstyle{rm}\Url}\fi

\bibitem[Baker et~al.(2023)Baker, Chen, and Ning]{baker2023rlsync}
Baker, F.~N., Chen, Z., and Ning, X.
\newblock Rlsync: Offline-online reinforcement learning for synthon completion.
\newblock \emph{arXiv preprint arXiv:2309.02671}, 2023.

\bibitem[Bhattacharyya et~al.(2020)Bhattacharyya, Rooshenas, Naskar, Sun, Iyyer, and McCallum]{bhattacharyya2020energy}
Bhattacharyya, S., Rooshenas, A., Naskar, S., Sun, S., Iyyer, M., and McCallum, A.
\newblock Energy-based reranking: Improving neural machine translation using energy-based models.
\newblock \emph{arXiv preprint arXiv:2009.13267}, 2020.

\bibitem[Bradley \& Terry(1952)Bradley and Terry]{bradley1952rank}
Bradley, R.~A. and Terry, M.~E.
\newblock Rank analysis of incomplete block designs: I. the method of paired comparisons.
\newblock \emph{Biometrika}, 39\penalty0 (3/4):\penalty0 324--345, 1952.

\bibitem[Brown et~al.(2020)Brown, Mann, Ryder, Subbiah, Kaplan, Dhariwal, Neelakantan, Shyam, Sastry, Askell, et~al.]{brown2020language}
Brown, T., Mann, B., Ryder, N., Subbiah, M., Kaplan, J.~D., Dhariwal, P., Neelakantan, A., Shyam, P., Sastry, G., Askell, A., et~al.
\newblock Language models are few-shot learners.
\newblock In \emph{Advances in Neural Information Processing Systems}, 2020.

\bibitem[Carreira-Perpinan \& Hinton(2005)Carreira-Perpinan and Hinton]{carreira2005contrastive}
Carreira-Perpinan, M.~A. and Hinton, G.
\newblock On contrastive divergence learning.
\newblock In \emph{International Workshop on Artificial Intelligence and Statistics}, 2005.

\bibitem[Chen et~al.(2019)Chen, Shen, Jaakkola, and Barzilay]{chen2019learning}
Chen, B., Shen, T., Jaakkola, T.~S., and Barzilay, R.
\newblock Learning to make generalizable and diverse predictions for retrosynthesis.
\newblock \emph{arXiv preprint arXiv:1910.09688}, 2019.

\bibitem[Chen et~al.(2020)Chen, Li, Dai, and Song]{chen2020retro}
Chen, B., Li, C., Dai, H., and Song, L.
\newblock Retro*: learning retrosynthetic planning with neural guided a* search.
\newblock In \emph{International Conference on Machine Learning}, 2020.

\bibitem[Chen \& Jung(2021)Chen and Jung]{chen2021deep}
Chen, S. and Jung, Y.
\newblock Deep retrosynthetic reaction prediction using local reactivity and global attention.
\newblock \emph{JACS Au}, 1\penalty0 (10):\penalty0 1612--1620, 2021.

\bibitem[Chen et~al.(2023)Chen, Ayinde, Fuchs, Sun, and Ning]{chen2023g2retro}
Chen, Z., Ayinde, O.~R., Fuchs, J.~R., Sun, H., and Ning, X.
\newblock G2retro as a two-step graph generative models for retrosynthesis prediction.
\newblock \emph{Communications Chemistry}, 6\penalty0 (1):\penalty0 102, 2023.

\bibitem[Coley et~al.(2017)Coley, Rogers, Green, and Jensen]{coley2017computer}
Coley, C.~W., Rogers, L., Green, W.~H., and Jensen, K.~F.
\newblock Computer-assisted retrosynthesis based on molecular similarity.
\newblock \emph{ACS Central Science}, 3\penalty0 (12):\penalty0 1237--1245, 2017.

\bibitem[Corey(1991)]{corey1991logic}
Corey, E.~J.
\newblock The logic of chemical synthesis: multistep synthesis of complex carbogenic molecules (nobel lecture).
\newblock \emph{Angewandte Chemie International Edition in English}, 30\penalty0 (5):\penalty0 455--465, 1991.

\bibitem[Corey \& Wipke(1969)Corey and Wipke]{corey1969computer}
Corey, E.~J. and Wipke, W.~T.
\newblock Computer-assisted design of complex organic syntheses: Pathways for molecular synthesis can be devised with a computer and equipment for graphical communication.
\newblock \emph{Science}, 166\penalty0 (3902):\penalty0 178--192, 1969.

\bibitem[Dai et~al.(2019)Dai, Li, Coley, Dai, and Song]{dai2019retrosynthesis}
Dai, H., Li, C., Coley, C., Dai, B., and Song, L.
\newblock Retrosynthesis prediction with conditional graph logic network.
\newblock In \emph{Advances in Neural Information Processing Systems}, 2019.

\bibitem[Deng et~al.(2020)Deng, Bakhtin, Ott, Szlam, and Ranzato]{deng2020residual}
Deng, Y., Bakhtin, A., Ott, M., Szlam, A., and Ranzato, M.
\newblock Residual energy-based models for text generation.
\newblock In \emph{International Conference on Learning Representations}, 2020.

\bibitem[Du \& Mordatch(2019)Du and Mordatch]{du2019implicit}
Du, Y. and Mordatch, I.
\newblock Implicit generation and generalization in energy-based models.
\newblock \emph{arXiv preprint arXiv:1903.08689}, 2019.

\bibitem[Du et~al.(2020)Du, Li, and Mordatch]{du2020compositional}
Du, Y., Li, S., and Mordatch, I.
\newblock Compositional visual generation with energy based models.
\newblock In \emph{Advances in Neural Information Processing Systems}, 2020.

\bibitem[Du et~al.(2021)Du, Li, Sharma, Tenenbaum, and Mordatch]{du2021unsupervised}
Du, Y., Li, S., Sharma, Y., Tenenbaum, J., and Mordatch, I.
\newblock Unsupervised learning of compositional energy concepts.
\newblock In \emph{Advances in Neural Information Processing Systems}, 2021.

\bibitem[Du et~al.(2023)Du, Durkan, Strudel, Tenenbaum, Dieleman, Fergus, Sohl-Dickstein, Doucet, and Grathwohl]{du2023reduce}
Du, Y., Durkan, C., Strudel, R., Tenenbaum, J.~B., Dieleman, S., Fergus, R., Sohl-Dickstein, J., Doucet, A., and Grathwohl, W.~S.
\newblock Reduce, reuse, recycle: Compositional generation with energy-based diffusion models and mcmc.
\newblock In \emph{International Conference on Machine Learning}, 2023.

\bibitem[Fang et~al.(2022)Fang, Li, Zhao, Tan, and Lou]{fang2022leveraging}
Fang, L., Li, J., Zhao, M., Tan, L., and Lou, J.-G.
\newblock Leveraging reaction-aware substructures for retrosynthesis analysis.
\newblock \emph{arXiv preprint arXiv:2204.05919}, 2022.

\bibitem[Gao et~al.(2022)Gao, Tan, Wu, and Li]{gao2022semiretro}
Gao, Z., Tan, C., Wu, L., and Li, S.~Z.
\newblock Semiretro: Semi-template framework boosts deep retrosynthesis prediction.
\newblock \emph{arXiv preprint arXiv:2202.08205}, 2022.

\bibitem[Gao et~al.(2023)Gao, Chen, Tan, and Li]{gao2023motifretro}
Gao, Z., Chen, X., Tan, C., and Li, S.~Z.
\newblock Motifretro: Exploring the combinability-consistency trade-offs in retrosynthesis via dynamic motif editing.
\newblock \emph{arXiv preprint arXiv:2305.15153}, 2023.

\bibitem[Genheden \& Bjerrum(2022)Genheden and Bjerrum]{genheden2022paroutes}
Genheden, S. and Bjerrum, E.
\newblock Paroutes: towards a framework for benchmarking retrosynthesis route predictions.
\newblock \emph{Digital Discovery}, 1\penalty0 (4):\penalty0 527--539, 2022.

\bibitem[Grathwohl et~al.(2019)Grathwohl, Wang, Jacobsen, Duvenaud, Norouzi, and Swersky]{grathwohl2019your}
Grathwohl, W., Wang, K.-C., Jacobsen, J.-H., Duvenaud, D., Norouzi, M., and Swersky, K.
\newblock Your classifier is secretly an energy based model and you should treat it like one.
\newblock \emph{arXiv preprint arXiv:1912.03263}, 2019.

\bibitem[Guo et~al.(2020)Guo, Wu, Ohno, and Yoshida]{guo2020bayesian}
Guo, Z., Wu, S., Ohno, M., and Yoshida, R.
\newblock Bayesian algorithm for retrosynthesis.
\newblock \emph{Journal of Chemical Information and Modeling}, 60\penalty0 (10):\penalty0 4474--4486, 2020.

\bibitem[Gutmann \& Hyv{\"a}rinen(2010)Gutmann and Hyv{\"a}rinen]{gutmann2010noise}
Gutmann, M. and Hyv{\"a}rinen, A.
\newblock Noise-contrastive estimation: A new estimation principle for unnormalized statistical models.
\newblock In \emph{International Conference on Artificial Intelligence and Statistics}, 2010.

\bibitem[Han et~al.(2022)Han, Zhao, Lu, Huang, Wu, Shang, Yao, and Zhang]{han2022gnn}
Han, P., Zhao, P., Lu, C., Huang, J., Wu, J., Shang, S., Yao, B., and Zhang, X.
\newblock Gnn-retro: Retrosynthetic planning with graph neural networks.
\newblock In \emph{Proceedings of the AAAI Conference on Artificial Intelligence}, 2022.

\bibitem[Haroutunian et~al.(2023)Haroutunian, Li, Galescu, Cohen, Tumuluri, and Haffari]{haroutunian2023reranking}
Haroutunian, L., Li, Z., Galescu, L., Cohen, P., Tumuluri, R., and Haffari, G.
\newblock Reranking for natural language generation from logical forms: A study based on large language models.
\newblock \emph{arXiv preprint arXiv:2309.12294}, 2023.

\bibitem[Hassen et~al.(2022)Hassen, Torren-Peraire, Genheden, Verhoeven, Preuss, and Tetko]{hassen2022mind}
Hassen, A.~K., Torren-Peraire, P., Genheden, S., Verhoeven, J., Preuss, M., and Tetko, I.
\newblock Mind the retrosynthesis gap: Bridging the divide between single-step and multi-step retrosynthesis prediction.
\newblock \emph{arXiv preprint arXiv:2212.11809}, 2022.

\bibitem[He et~al.(2022)He, Wang, Liu, and Wu]{he2022modeling}
He, H.-R., Wang, J., Liu, Y., and Wu, F.
\newblock Modeling diverse chemical reactions for single-step retrosynthesis via discrete latent variables.
\newblock In \emph{Proceedings of the 31st ACM International Conference on Information \& Knowledge Management}, 2022.

\bibitem[Heifets \& Jurisica(2012)Heifets and Jurisica]{heifets2012construction}
Heifets, A. and Jurisica, I.
\newblock Construction of new medicines via game proof search.
\newblock In \emph{Proceedings of the AAAI Conference on Artificial Intelligence}, 2012.

\bibitem[Hinton(2002)]{hinton2002training}
Hinton, G.~E.
\newblock Training products of experts by minimizing contrastive divergence.
\newblock \emph{Neural computation}, 14\penalty0 (8):\penalty0 1771--1800, 2002.

\bibitem[Hoffmann(2009)]{hoffmann2009elements}
Hoffmann, R.~W.
\newblock \emph{Elements of synthesis planning}, volume 307.
\newblock Springer, 2009.

\bibitem[Hong et~al.(2021)Hong, Zhuo, Jin, and Zhou]{hong2021retrosynthetic}
Hong, S., Zhuo, H.~H., Jin, K., and Zhou, Z.
\newblock Retrosynthetic planning with experience-guided monte carlo tree search.
\newblock \emph{arXiv preprint arXiv:2112.06028}, 2021.

\bibitem[Igashov et~al.(2024)Igashov, Schneuing, Segler, Bronstein, and Correia]{igashov2024retrobridge}
Igashov, I., Schneuing, A., Segler, M., Bronstein, M., and Correia, B.
\newblock Retrobridge: Modeling retrosynthesis with markov bridges.
\newblock In \emph{International Conference on Learning Representations}, 2024.

\bibitem[Ishiguro et~al.(2020)Ishiguro, Ujihara, Sawada, Akita, and Kotera]{ishiguro2020data}
Ishiguro, K., Ujihara, K., Sawada, R., Akita, H., and Kotera, M.
\newblock Data transfer approaches to improve seq-to-seq retrosynthesis.
\newblock \emph{arXiv preprint arXiv:2010.00792}, 2020.

\bibitem[Jiang et~al.(2023)Jiang, Ying, Wu, Huang, Kuang, and Wang]{jiang2023learning}
Jiang, Y., Ying, W., Wu, F., Huang, Z., Kuang, K., and Wang, Z.
\newblock Learning chemical rules of retrosynthesis with pre-training.
\newblock In \emph{Proceedings of the AAAI Conference on Artificial Intelligence}, 2023.

\bibitem[Karpov et~al.(2019)Karpov, Godin, and Tetko]{karpov2019transformer}
Karpov, P., Godin, G., and Tetko, I.~V.
\newblock A transformer model for retrosynthesis.
\newblock In \emph{International Conference on Artificial Neural Networks}, 2019.

\bibitem[Kim et~al.(2021)Kim, Ahn, Lee, and Shin]{kim2021self}
Kim, J., Ahn, S., Lee, H., and Shin, J.
\newblock Self-improved retrosynthetic planning.
\newblock In \emph{International Conference on Machine Learning}, 2021.

\bibitem[Kishimoto et~al.(2019)Kishimoto, Buesser, Chen, and Botea]{kishimoto2019depth}
Kishimoto, A., Buesser, B., Chen, B., and Botea, A.
\newblock Depth-first proof-number search with heuristic edge cost and application to chemical synthesis planning.
\newblock In \emph{Advances in Neural Information Processing Systems}, 2019.

\bibitem[Krause et~al.(2020)Krause, Gotmare, McCann, Keskar, Joty, Socher, and Rajani]{krause2020gedi}
Krause, B., Gotmare, A.~D., McCann, B., Keskar, N.~S., Joty, S., Socher, R., and Rajani, N.~F.
\newblock Gedi: Generative discriminator guided sequence generation.
\newblock \emph{arXiv preprint arXiv:2009.06367}, 2020.

\bibitem[Lan et~al.(2023)Lan, Zeng, Hong, Liu, and Ma]{lan2023rcsearcher}
Lan, Z., Zeng, Z., Hong, B., Liu, Z., and Ma, F.
\newblock Rcsearcher: Reaction center identification in retrosynthesis via deep q-learning.
\newblock \emph{arXiv preprint arXiv:2301.12071}, 2023.

\bibitem[Lan et~al.(2024)Lan, Hong, Zhu, Zeng, Liu, Yu, and Ma]{lan2024retrosynthesis}
Lan, Z., Hong, B., Zhu, J., Zeng, Z., Liu, Z., Yu, L., and Ma, F.
\newblock Retrosynthesis prediction via search in (hyper) graph.
\newblock \emph{arXiv preprint arXiv:2402.06772}, 2024.

\bibitem[LeCun et~al.(2006)LeCun, Chopra, Hadsell, Ranzato, and Huang]{lecun2006tutorial}
LeCun, Y., Chopra, S., Hadsell, R., Ranzato, M., and Huang, F.
\newblock A tutorial on energy-based learning.
\newblock \emph{Predicting structured data}, 1\penalty0 (0), 2006.

\bibitem[Lee et~al.(2021)Lee, Ahn, Seo, Song, Yang, Hwang, and Shin]{lee2021retcl}
Lee, H., Ahn, S., Seo, S.-W., Song, Y.~Y., Yang, E., Hwang, S.-J., and Shin, J.
\newblock Retcl: A selection-based approach for retrosynthesis via contrastive learning.
\newblock \emph{arXiv preprint arXiv:2105.00795}, 2021.

\bibitem[Lee et~al.(2023)Lee, Kim, Choi, Kwak, Park, Hwang, and Kim]{lee2023readretro}
Lee, S., Kim, T., Choi, M.-S., Kwak, Y., Park, J., Hwang, S.~J., and Kim, S.-G.
\newblock Readretro: Natural product biosynthesis planning with retrieval-augmented dual-view retrosynthesis.
\newblock \emph{bioRxiv}, pp.\  2023--03, 2023.

\bibitem[Li et~al.(2023{\natexlab{a}})Li, Fang, and Lou]{li2023retro}
Li, J., Fang, L., and Lou, J.-G.
\newblock Retro-bleu: Quantifying chemical plausibility of retrosynthesis routes through reaction template sequence analysis.
\newblock \emph{arXiv preprint arXiv:2311.06304}, 2023{\natexlab{a}}.

\bibitem[Li et~al.(2023{\natexlab{b}})Li, Fang, and Lou]{li2023retroranker}
Li, J., Fang, L., and Lou, J.-G.
\newblock Retroranker: leveraging reaction changes to improve retrosynthesis prediction through re-ranking.
\newblock \emph{Journal of Cheminformatics}, 15\penalty0 (1):\penalty0 58, 2023{\natexlab{b}}.

\bibitem[Li et~al.(2022)Li, Thickstun, Gulrajani, Liang, and Hashimoto]{li2022diffusion}
Li, X., Thickstun, J., Gulrajani, I., Liang, P.~S., and Hashimoto, T.~B.
\newblock Diffusion-lm improves controllable text generation.
\newblock In \emph{Advances in Neural Information Processing Systems}, 2022.

\bibitem[Lin et~al.(2022)Lin, Tu, and Coley]{lin2022improving}
Lin, M.~H., Tu, Z., and Coley, C.~W.
\newblock Improving the performance of models for one-step retrosynthesis through re-ranking.
\newblock \emph{Journal of cheminformatics}, 14\penalty0 (1):\penalty0 1--13, 2022.

\bibitem[Lin et~al.(2023)Lin, Yin, Shi, Zhou, and Zhang]{lin2023g2gt}
Lin, Z., Yin, S., Shi, L., Zhou, W., and Zhang, Y.~J.
\newblock G2gt: Retrosynthesis prediction with graph-to-graph attention neural network and self-training.
\newblock \emph{Journal of Chemical Information and Modeling}, 63\penalty0 (7):\penalty0 1894--1905, 2023.

\bibitem[Liu et~al.(2021)Liu, Sap, Lu, Swayamdipta, Bhagavatula, Smith, and Choi]{liu2021dexperts}
Liu, A., Sap, M., Lu, X., Swayamdipta, S., Bhagavatula, C., Smith, N.~A., and Choi, Y.
\newblock Dexperts: Decoding-time controlled text generation with experts and anti-experts.
\newblock \emph{arXiv preprint arXiv:2105.03023}, 2021.

\bibitem[Liu et~al.(2017)Liu, Ramsundar, Kawthekar, Shi, Gomes, Luu~Nguyen, Ho, Sloane, Wender, and Pande]{liu2017retrosynthetic}
Liu, B., Ramsundar, B., Kawthekar, P., Shi, J., Gomes, J., Luu~Nguyen, Q., Ho, S., Sloane, J., Wender, P., and Pande, V.
\newblock Retrosynthetic reaction prediction using neural sequence-to-sequence models.
\newblock \emph{ACS Central Science}, 3\penalty0 (10):\penalty0 1103--1113, 2017.

\bibitem[Liu et~al.(2023{\natexlab{a}})Liu, Xue, Xie, Xia, Tripp, Maziarz, Segler, Qin, Zhang, and Liu]{liu2023retrosynthetic}
Liu, G., Xue, D., Xie, S., Xia, Y., Tripp, A., Maziarz, K., Segler, M., Qin, T., Zhang, Z., and Liu, T.-Y.
\newblock Retrosynthetic planning with dual value networks.
\newblock In \emph{International Conference on Machine Learning}, 2023{\natexlab{a}}.

\bibitem[Liu et~al.(2022)Liu, Yan, Yu, Lu, Huang, Ou-Yang, and Zhao]{liu2022mars}
Liu, J., Yan, C., Yu, Y., Lu, C., Huang, J., Ou-Yang, L., and Zhao, P.
\newblock Mars: A motif-based autoregressive model for retrosynthesis prediction.
\newblock \emph{arXiv preprint arXiv:2209.13178}, 2022.

\bibitem[Liu et~al.(2023{\natexlab{b}})Liu, Tu, Xu, Zhang, Lin, Ying, Tang, Zhao, and Wu]{liu2023fusionretro}
Liu, S., Tu, Z., Xu, M., Zhang, Z., Lin, L., Ying, R., Tang, J., Zhao, P., and Wu, D.
\newblock Fusionretro: Molecule representation fusion via in-context learning for retrosynthetic planning.
\newblock In \emph{International Conference on Machine Learning}, 2023{\natexlab{b}}.

\bibitem[Liu et~al.(2024)Liu, Xu, Fang, Xi, Liu, Zhang, Poon, and Wang]{liu2024t}
Liu, Y., Xu, H., Fang, T., Xi, H., Liu, Z., Zhang, S., Poon, H., and Wang, S.
\newblock T-rex: Text-assisted retrosynthesis prediction.
\newblock \emph{arXiv preprint arXiv:2401.14637}, 2024.

\bibitem[Maziarz et~al.(2023)Maziarz, Tripp, Liu, Stanley, Xie, Gai{\'n}ski, Seidl, and Segler]{maziarz2023re}
Maziarz, K., Tripp, A., Liu, G., Stanley, M., Xie, S., Gai{\'n}ski, P., Seidl, P., and Segler, M.
\newblock Re-evaluating retrosynthesis algorithms with syntheseus.
\newblock \emph{arXiv preprint arXiv:2310.19796}, 2023.

\bibitem[Meng et~al.(2023)Meng, Zhao, Yu, and King]{meng2023retro}
Meng, Z., Zhao, P., Yu, Y., and King, I.
\newblock A unified view of deep learning for reaction and retrosynthesis prediction: Current status and future challenges.
\newblock In \emph{Proceedings of the International Joint Conference on Artificial Intelligence}, 2023.

\bibitem[Ouyang et~al.(2022)Ouyang, Wu, Jiang, Almeida, Wainwright, Mishkin, Zhang, Agarwal, Slama, Ray, et~al.]{ouyang2022training}
Ouyang, L., Wu, J., Jiang, X., Almeida, D., Wainwright, C., Mishkin, P., Zhang, C., Agarwal, S., Slama, K., Ray, A., et~al.
\newblock Training language models to follow instructions with human feedback.
\newblock In \emph{Advances in Neural Information Processing Systems}, 2022.

\bibitem[Parshakova et~al.(2019)Parshakova, Andreoli, and Dymetman]{parshakova2019global}
Parshakova, T., Andreoli, J.-M., and Dymetman, M.
\newblock Global autoregressive models for data-efficient sequence learning.
\newblock \emph{arXiv preprint arXiv:1909.07063}, 2019.

\bibitem[Paszke et~al.(2019)Paszke, Gross, Massa, Lerer, Bradbury, Chanan, Killeen, Lin, Gimelshein, Antiga, et~al.]{paszke2019pytorch}
Paszke, A., Gross, S., Massa, F., Lerer, A., Bradbury, J., Chanan, G., Killeen, T., Lin, Z., Gimelshein, N., Antiga, L., et~al.
\newblock Pytorch: An imperative style, high-performance deep learning library.
\newblock In \emph{Advances in Neural Information Processing Systems}, 2019.

\bibitem[Qian et~al.(2023)Qian, Li, Tu, Coley, and Barzilay]{qian2023predictive}
Qian, Y., Li, Z., Tu, Z., Coley, C.~W., and Barzilay, R.
\newblock Predictive chemistry augmented with text retrieval.
\newblock \emph{arXiv preprint arXiv:2312.04881}, 2023.

\bibitem[Rafailov et~al.(2023)Rafailov, Sharma, Mitchell, Ermon, Manning, and Finn]{rafailov2023direct}
Rafailov, R., Sharma, A., Mitchell, E., Ermon, S., Manning, C.~D., and Finn, C.
\newblock Direct preference optimization: Your language model is secretly a reward model.
\newblock In \emph{Advances in Neural Information Processing Systems}, 2023.

\bibitem[Ranzato et~al.(2007)Ranzato, Boureau, Chopra, and LeCun]{ranzato2007unified}
Ranzato, M., Boureau, Y.-L., Chopra, S., and LeCun, Y.
\newblock A unified energy-based framework for unsupervised learning.
\newblock In \emph{International Conference on Artificial Intelligence and Statistics}, 2007.

\bibitem[Ranzato et~al.(2016)Ranzato, Chopra, Auli, and Zaremba]{ranzato2016sequence}
Ranzato, M., Chopra, S., Auli, M., and Zaremba, W.
\newblock Sequence level training with recurrent neural networks.
\newblock In \emph{International Conference on Learning Representations}, 2016.

\bibitem[Sacha et~al.(2021)Sacha, B{\l}az, Byrski, Dabrowski-Tumanski, Chrominski, Loska, W{\l}odarczyk-Pruszynski, and Jastrzebski]{sacha2021molecule}
Sacha, M., B{\l}az, M., Byrski, P., Dabrowski-Tumanski, P., Chrominski, M., Loska, R., W{\l}odarczyk-Pruszynski, P., and Jastrzebski, S.
\newblock Molecule edit graph attention network: modeling chemical reactions as sequences of graph edits.
\newblock \emph{Journal of Chemical Information and Modeling}, 61\penalty0 (7):\penalty0 3273--3284, 2021.

\bibitem[Sacha et~al.(2023)Sacha, Sadowski, Kozakowski, van Workum, and Jastrzebski]{sacha2023molecule}
Sacha, M., Sadowski, M., Kozakowski, P., van Workum, R., and Jastrzebski, S.
\newblock Molecule-edit templates for efficient and accurate retrosynthesis prediction.
\newblock \emph{arXiv preprint arXiv:2310.07313}, 2023.

\bibitem[Segler \& Waller(2017)Segler and Waller]{segler2017neural}
Segler, M.~H. and Waller, M.~P.
\newblock Neural-symbolic machine learning for retrosynthesis and reaction prediction.
\newblock \emph{Chemistry--A European Journal}, 23\penalty0 (25):\penalty0 5966--5971, 2017.

\bibitem[Segler et~al.(2018)Segler, Preuss, and Waller]{segler2018planning}
Segler, M.~H., Preuss, M., and Waller, M.~P.
\newblock Planning chemical syntheses with deep neural networks and symbolic ai.
\newblock \emph{Nature}, 555\penalty0 (7698):\penalty0 604, 2018.

\bibitem[Seidl et~al.(2021)Seidl, Renz, Dyubankova, Neves, Verhoeven, Wegner, Hochreiter, and Klambauer]{seidl2021modern}
Seidl, P., Renz, P., Dyubankova, N., Neves, P., Verhoeven, J., Wegner, J.~K., Hochreiter, S., and Klambauer, G.
\newblock Modern hopfield networks for few-and zero-shot reaction prediction.
\newblock \emph{arXiv preprint arXiv:2104.03279}, 2021.

\bibitem[Seo et~al.(2021)Seo, Song, Yang, Bae, Lee, Shin, Hwang, and Yang]{seo2021gta}
Seo, S.-W., Song, Y.~Y., Yang, J.~Y., Bae, S., Lee, H., Shin, J., Hwang, S.~J., and Yang, E.
\newblock Gta: Graph truncated attention for retrosynthesis.
\newblock In \emph{Proceedings of the AAAI Conference on Artificial Intelligence}, 2021.

\bibitem[Shi et~al.(2020)Shi, Xu, Guo, Zhang, and Tang]{shi2020graph}
Shi, C., Xu, M., Guo, H., Zhang, M., and Tang, J.
\newblock A graph to graphs framework for retrosynthesis prediction.
\newblock In \emph{International Conference on Machine Learning}, 2020.

\bibitem[Somnath et~al.(2021)Somnath, Bunne, Coley, Krause, and Barzilay]{somnath2021learning}
Somnath, V.~R., Bunne, C., Coley, C., Krause, A., and Barzilay, R.
\newblock Learning graph models for retrosynthesis prediction.
\newblock In \emph{Advances in Neural Information Processing Systems}, 2021.

\bibitem[Song \& Kingma(2021)Song and Kingma]{song2021train}
Song, Y. and Kingma, D.~P.
\newblock How to train your energy-based models.
\newblock \emph{arXiv preprint arXiv:2101.03288}, 2021.

\bibitem[Sun et~al.(2023)Sun, Dai, Dai, Zhou, and Schuurmans]{sun2023discrete}
Sun, H., Dai, H., Dai, B., Zhou, H., and Schuurmans, D.
\newblock Discrete langevin samplers via wasserstein gradient flow.
\newblock In \emph{International Conference on Artificial Intelligence and Statistics}, 2023.

\bibitem[Sun et~al.(2021)Sun, Dai, Li, Kearnes, and Dai]{sun2021towards}
Sun, R., Dai, H., Li, L., Kearnes, S., and Dai, B.
\newblock Towards understanding retrosynthesis by energy-based models.
\newblock In \emph{Advances in Neural Information Processing Systems}, 2021.

\bibitem[Tetko et~al.(2020)Tetko, Karpov, Van~Deursen, and Godin]{tetko2020state}
Tetko, I.~V., Karpov, P., Van~Deursen, R., and Godin, G.
\newblock State-of-the-art augmented nlp transformer models for direct and single-step retrosynthesis.
\newblock \emph{Nature communications}, 11\penalty0 (1):\penalty0 1--11, 2020.

\bibitem[Torren-Peraire et~al.(2023)Torren-Peraire, Hassen, Genheden, Verhoeven, Clevert, Preuss, and Tetko]{torren2023models}
Torren-Peraire, P., Hassen, A.~K., Genheden, S., Verhoeven, J., Clevert, D.-A., Preuss, M., and Tetko, I.
\newblock Models matter: The impact of single-step retrosynthesis on synthesis planning.
\newblock \emph{arXiv preprint arXiv:2308.05522}, 2023.

\bibitem[Touvron et~al.(2023)Touvron, Lavril, Izacard, Martinet, Lachaux, Lacroix, Rozi{\`e}re, Goyal, Hambro, Azhar, et~al.]{touvron2023llama}
Touvron, H., Lavril, T., Izacard, G., Martinet, X., Lachaux, M.-A., Lacroix, T., Rozi{\`e}re, B., Goyal, N., Hambro, E., Azhar, F., et~al.
\newblock Llama: Open and efficient foundation language models.
\newblock \emph{arXiv preprint arXiv:2302.13971}, 2023.

\bibitem[Tripp et~al.(2022)Tripp, Maziarz, Lewis, Liu, and Segler]{tripp2022re}
Tripp, A., Maziarz, K., Lewis, S., Liu, G., and Segler, M.
\newblock Re-evaluating chemical synthesis planning algorithms.
\newblock In \emph{NeurIPS 2022 AI for Science: Progress and Promises}, 2022.

\bibitem[Tripp et~al.(2024)Tripp, Maziarz, Lewis, Segler, and Hern{\'a}ndez-Lobato]{tripp2024retro}
Tripp, A., Maziarz, K., Lewis, S., Segler, M., and Hern{\'a}ndez-Lobato, J.~M.
\newblock Retro-fallback: retrosynthetic planning in an uncertain world.
\newblock In \emph{International Conference on Learning Representations}, 2024.

\bibitem[Tu et~al.(2022)Tu, Shorewala, Ma, and Thost]{tu2022retrosynthesis}
Tu, H., Shorewala, S., Ma, T., and Thost, V.
\newblock Retrosynthesis prediction revisited.
\newblock In \emph{NeurIPS 2022 AI for Science: Progress and Promises}, 2022.

\bibitem[Tu \& Coley(2022)Tu and Coley]{tu2022permutation}
Tu, Z. and Coley, C.~W.
\newblock Permutation invariant graph-to-sequence model for template-free retrosynthesis and reaction prediction.
\newblock \emph{Journal of chemical information and modeling}, 62\penalty0 (15):\penalty0 3503--3513, 2022.

\bibitem[Variani et~al.(2022)Variani, Wu, Riley, Rybach, Shannon, and Allauzen]{variani2022global}
Variani, E., Wu, K., Riley, M.~D., Rybach, D., Shannon, M., and Allauzen, C.
\newblock Global normalization for streaming speech recognition in a modular framework.
\newblock In \emph{Advances in Neural Information Processing Systems}, 2022.

\bibitem[Vaswani et~al.(2017)Vaswani, Shazeer, Parmar, Uszkoreit, Jones, Gomez, Kaiser, and Polosukhin]{vaswani2017attention}
Vaswani, A., Shazeer, N., Parmar, N., Uszkoreit, J., Jones, L., Gomez, A.~N., Kaiser, {\L}., and Polosukhin, I.
\newblock Attention is all you need.
\newblock In \emph{Advances in Neural Information Processing Systems}, 2017.

\bibitem[Wan et~al.(2022)Wan, Hsieh, Liao, and Zhang]{wan2022retroformer}
Wan, Y., Hsieh, C.-Y., Liao, B., and Zhang, S.
\newblock Retroformer: Pushing the limits of end-to-end retrosynthesis transformer.
\newblock In \emph{International Conference on Machine Learning}, 2022.

\bibitem[Wang \& Ou(2018)Wang and Ou]{wang2018learning}
Wang, B. and Ou, Z.
\newblock Learning neural trans-dimensional random field language models with noise-contrastive estimation.
\newblock In \emph{2018 IEEE International Conference on Acoustics, Speech and Signal Processing}, 2018.

\bibitem[Wang et~al.(2023)Wang, Song, Xu, Wang, Zhou, and Ma]{wang2023retrodiff}
Wang, Y., Song, Y., Xu, M., Wang, R., Zhou, H., and Ma, W.
\newblock Retrodiff: Retrosynthesis as multi-stage distribution interpolation.
\newblock \emph{arXiv preprint arXiv:2311.14077}, 2023.

\bibitem[Xie et~al.(2016)Xie, Lu, Zhu, and Wu]{xie2016theory}
Xie, J., Lu, Y., Zhu, S.-C., and Wu, Y.
\newblock A theory of generative convnet.
\newblock In \emph{International Conference on Machine Learning}, 2016.

\bibitem[Xie et~al.(2022)Xie, Yan, Han, Xia, Wu, Guo, Yang, and Qin]{xie2022retrograph}
Xie, S., Yan, R., Han, P., Xia, Y., Wu, L., Guo, C., Yang, B., and Qin, T.
\newblock Retrograph: Retrosynthetic planning with graph search.
\newblock In \emph{Proceedings of the 28th ACM SIGKDD Conference on Knowledge Discovery and Data Mining}, 2022.

\bibitem[Xie et~al.(2023)Xie, Yan, Guo, Xia, Wu, and Qin]{xie2023retrosynthesis}
Xie, S., Yan, R., Guo, J., Xia, Y., Wu, L., and Qin, T.
\newblock Retrosynthesis prediction with local template retrieval.
\newblock \emph{arXiv preprint arXiv:2306.04123}, 2023.

\bibitem[Xiong et~al.(2023)Xiong, Zhang, Fu, Huang, Kong, Wang, Xiong, and Zheng]{xiong2023improve}
Xiong, J., Zhang, W., Fu, Z., Huang, J., Kong, X., Wang, Y., Xiong, Z., and Zheng, M.
\newblock Improve retrosynthesis planning with a molecular editing language.
\newblock \emph{chemrxiv}, 2023.

\bibitem[Yan et~al.(2020)Yan, Ding, Zhao, Zheng, Yang, Yu, and Huang]{yan2020retroxpert}
Yan, C., Ding, Q., Zhao, P., Zheng, S., Yang, J., Yu, Y., and Huang, J.
\newblock Retroxpert: Decompose retrosynthesis prediction like a chemist.
\newblock In \emph{Advances in Neural Information Processing Systems}, 2020.

\bibitem[Yan et~al.(2022)Yan, Zhao, Lu, Yu, and Huang]{yan2022retrocomposer}
Yan, C., Zhao, P., Lu, C., Yu, Y., and Huang, J.
\newblock Retrocomposer: Composing templates for template-based retrosynthesis prediction.
\newblock \emph{Biomolecules}, 12\penalty0 (9):\penalty0 1325, 2022.

\bibitem[Yang \& Klein(2021)Yang and Klein]{yang2021fudge}
Yang, K. and Klein, D.
\newblock Fudge: Controlled text generation with future discriminators.
\newblock \emph{arXiv preprint arXiv:2104.05218}, 2021.

\bibitem[Yao et~al.(2023)Yao, Wang, Guo, Xiang, Liu, and Ke]{yao2023node}
Yao, L., Wang, Z., Guo, W., Xiang, S., Liu, W., and Ke, G.
\newblock Node-aligned graph-to-graph generation for retrosynthesis prediction.
\newblock \emph{arXiv preprint arXiv:2309.15798}, 2023.

\bibitem[Yu et~al.(2022)Yu, Wei, Kuang, Huang, Yao, and Wu]{yu2022grasp}
Yu, Y., Wei, Y., Kuang, K., Huang, Z., Yao, H., and Wu, F.
\newblock Grasp: Navigating retrosynthetic planning with goal-driven policy.
\newblock In \emph{Advances in Neural Information Processing Systems}, 2022.

\bibitem[Yu et~al.(2023)Yu, Yuan, Wei, Gao, Ye, Wang, and Wu]{yu2023retroood}
Yu, Y., Yuan, L., Wei, Y., Gao, H., Ye, X., Wang, Z., and Wu, F.
\newblock Retroood: Understanding out-of-distribution generalization in retrosynthesis prediction.
\newblock \emph{arXiv preprint arXiv:2312.10900}, 2023.

\bibitem[Yuan et~al.(2024)Yuan, Yu, Wei, Wang, Wang, and Wu]{yuan2024active}
Yuan, L., Yu, Y., Wei, Y., Wang, Y., Wang, Z., and Wu, F.
\newblock Active retrosynthetic planning aware of route quality.
\newblock In \emph{International Conference on Learning Representations}, 2024.

\bibitem[Zhang et~al.(2023{\natexlab{a}})Zhang, Rao, and Agrawala]{zhang2023adding}
Zhang, L., Rao, A., and Agrawala, M.
\newblock Adding conditional control to text-to-image diffusion models.
\newblock In \emph{Proceedings of the IEEE/CVF International Conference on Computer Vision}, 2023{\natexlab{a}}.

\bibitem[Zhang et~al.(2024{\natexlab{a}})Zhang, Liu, Zhang, Yang, Yang, and Zhang]{zhang2024multi}
Zhang, Q., Liu, J., Zhang, W., Yang, F., Yang, Z., and Zhang, X.
\newblock A multi-stream network for retrosynthesis prediction.
\newblock \emph{Frontiers of Computer Science}, 18\penalty0 (2):\penalty0 182906, 2024{\natexlab{a}}.

\bibitem[Zhang et~al.(2024{\natexlab{b}})Zhang, Mo, Wang, and Yang]{zhang2024retrosynthesis}
Zhang, X., Mo, Y., Wang, W., and Yang, Y.
\newblock Retrosynthesis prediction enhanced by in-silico reaction data augmentation.
\newblock \emph{arXiv preprint arXiv:2402.00086}, 2024{\natexlab{b}}.

\bibitem[Zhang et~al.(2023{\natexlab{b}})Zhang, Hao, He, Gao, and Zhou]{zhang2023evolutionary}
Zhang, Y., Hao, H., He, X., Gao, S., and Zhou, A.
\newblock Evolutionary retrosynthetic route planning.
\newblock \emph{arXiv preprint arXiv:2310.05186}, 2023{\natexlab{b}}.

\bibitem[Zheng et~al.(2019)Zheng, Rao, Zhang, Xu, and Yang]{zheng2019predicting}
Zheng, S., Rao, J., Zhang, Z., Xu, J., and Yang, Y.
\newblock Predicting retrosynthetic reactions using self-corrected transformer neural networks.
\newblock \emph{Journal of Chemical Information and Modeling}, 60\penalty0 (1):\penalty0 47--55, 2019.

\bibitem[Zhong et~al.(2023{\natexlab{a}})Zhong, Yang, and Chen]{zhong2023retrosynthesis}
Zhong, W., Yang, Z., and Chen, C. Y.-C.
\newblock Retrosynthesis prediction using an end-to-end graph generative architecture for molecular graph editing.
\newblock \emph{Nature Communications}, 14\penalty0 (1):\penalty0 3009, 2023{\natexlab{a}}.

\bibitem[Zhong et~al.(2022)Zhong, Song, Feng, Liu, Jia, Yao, Wu, Hou, and Song]{zhong2022root}
Zhong, Z., Song, J., Feng, Z., Liu, T., Jia, L., Yao, S., Wu, M., Hou, T., and Song, M.
\newblock Root-aligned smiles: a tight representation for chemical reaction prediction.
\newblock \emph{Chemical Science}, 13\penalty0 (31):\penalty0 9023--9034, 2022.

\bibitem[Zhong et~al.(2023{\natexlab{b}})Zhong, Song, Feng, Liu, Jia, Yao, Hou, and Song]{zhong2023recent}
Zhong, Z., Song, J., Feng, Z., Liu, T., Jia, L., Yao, S., Hou, T., and Song, M.
\newblock Recent advances in artificial intelligence for retrosynthesis.
\newblock \emph{arXiv preprint arXiv:2301.05864}, 2023{\natexlab{b}}.

\bibitem[Zhong et~al.(2024)Zhong, Song, Feng, Liu, Jia, Yao, Hou, and Song]{zhongrecent}
Zhong, Z., Song, J., Feng, Z., Liu, T., Jia, L., Yao, S., Hou, T., and Song, M.
\newblock Recent advances in deep learning for retrosynthesis.
\newblock \emph{Wiley Interdisciplinary Reviews: Computational Molecular Science}, pp.\  e1694, 2024.

\bibitem[Zhu et~al.(2023{\natexlab{a}})Zhu, Hong, Lan, and Ma]{zhu2023single}
Zhu, J., Hong, B., Lan, Z., and Ma, F.
\newblock Single-step retrosynthesis via reaction center and leaving groups prediction.
\newblock In \emph{2023 16th International Congress on Image and Signal Processing, BioMedical Engineering and Informatics}, 2023{\natexlab{a}}.

\bibitem[Zhu et~al.(2023{\natexlab{b}})Zhu, Xia, Wu, Xie, Zhou, Qin, Li, and Liu]{zhu2023dual}
Zhu, J., Xia, Y., Wu, L., Xie, S., Zhou, W., Qin, T., Li, H., and Liu, T.-Y.
\newblock Dual-view molecular pre-training.
\newblock In \emph{Proceedings of the 29th ACM SIGKDD Conference on Knowledge Discovery and Data Mining}, 2023{\natexlab{b}}.

\end{thebibliography}
\bibliographystyle{icml2024}

%%%%%%%%%%%%%%%%%%%%%%%%%%%%%%%%%%%%%%%%%%%%%%%%%%%%%%%%%%%%%%%%%%%%%%%%%%%%%%%
%%%%%%%%%%%%%%%%%%%%%%%%%%%%%%%%%%%%%%%%%%%%%%%%%%%%%%%%%%%%%%%%%%%%%%%%%%%%%%%
% APPENDIX
%%%%%%%%%%%%%%%%%%%%%%%%%%%%%%%%%%%%%%%%%%%%%%%%%%%%%%%%%%%%%%%%%%%%%%%%%%%%%%%
%%%%%%%%%%%%%%%%%%%%%%%%%%%%%%%%%%%%%%%%%%%%%%%%%%%%%%%%%%%%%%%%%%%%%%%%%%%%%%%
\newpage
\appendix
\onecolumn
\begin{figure}[t]
    \begin{center}
        \includegraphics[width=0.479\textwidth]{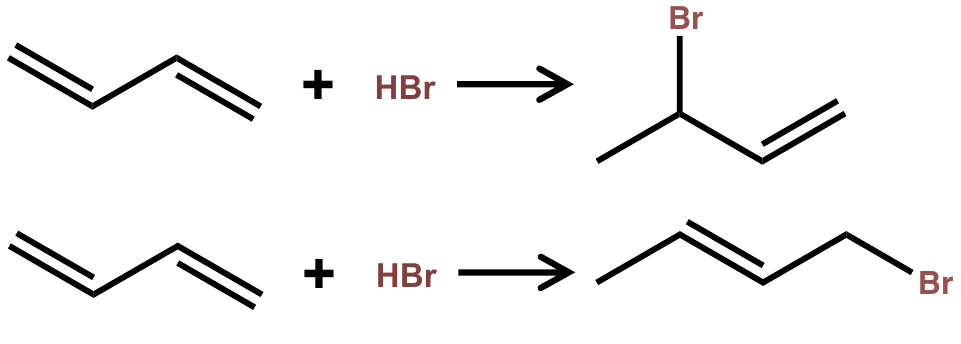}
    \end{center}
    \vskip -0.1in
    \caption{Hydrogen bromide (HBr) can undergo 1,2-addition and 1,4-addition with 1,3-butadiene under different reaction conditions.}
   \label{fig:12_14}
\end{figure}
\section{Discriminative Reaction Prediction vs. Generative Retrosynthesis Prediction}
\label{appendix:diff}
Reaction prediction and retrosynthetic prediction are two different types of tasks. With given reactants and reaction conditions, the output product is definite. Although the reactants can undergo 1-2 and 1-4 addition reactions to synthesize different products, as shown in ~\ref{fig:12_14}, the required reaction conditions are different. However, given a product molecule, the specific reaction conditions for its synthesis are not always known, and there could be multiple sets of reactants capable of synthesizing it.
\section{Related Work}
\label{appendix:related}
\paragraph{One-step Retrosynthesis Model.} Current one-step retrosynthesis models~\citep{meng2023retro,zhong2023recent,zhongrecent} fall into three distinct categories: template-based~\citep{coley2017computer,segler2017neural,dai2019retrosynthesis,chen2020retro,chen2021deep,seidl2021modern,yan2022retrocomposer,xie2023retrosynthesis,sacha2023molecule,zhang2024multi}, semi-template-based~\citep{shi2020graph,yan2020retroxpert,somnath2021learning,gao2022semiretro,zhong2023retrosynthesis,baker2023rlsync,wang2023retrodiff,gao2023motifretro,zhu2023single,lan2023rcsearcher,chen2023g2retro,lan2024retrosynthesis}, and template-free~\citep{liu2017retrosynthetic,zheng2019predicting,chen2019learning,karpov2019transformer,ishiguro2020data,guo2020bayesian,tetko2020state,seo2021gta,lee2021retcl,sun2021towards,fang2022leveraging,wan2022retroformer,he2022modeling,liu2022mars,tu2022permutation,lin2022improving,zhong2022root,yu2023retroood,li2023retroranker,zhu2023dual,jiang2023learning,qian2023predictive,xiong2023improve,yao2023node,lin2023g2gt,liu2024t,zhang2024retrosynthesis}. Template-based methods model retrosynthesis as either a classification or a template retrieval problem. These approaches involve extracting templates based on chemical rules, followed by training a model to identify the specific template that allows the product to be decomposed into reactants. During the inference stage, the process begins by predicting the class of the template rule, which is then applied to the product to deduce the corresponding reactants. Semi-template-based methods, often referred to as two-stage models, initially use templates to identify the reaction centers in a product. Subsequently, they break these centers down into several disconnected subgraphs, known as synthons. These synthons are then transformed into chemically valid molecules through generative modeling or by attaching leaving groups. Template-free approaches model retrosynthesis prediction as either a sequence-to-sequence task~\citep{karpov2019transformer}, using SMILES strings to represent molecules, or as a graph-edit problem~\citep{sacha2021molecule,igashov2024retrobridge}, where molecules are depicted as graphs.
\paragraph{Search Algorithm in Retrosynthetic Planning.} 
Search algorithms~\citep{segler2018planning,hong2021retrosynthetic,kishimoto2019depth,chen2020retro,heifets2012construction,kim2021self,yu2022grasp,han2022gnn,hassen2022mind,li2023retro,zhang2023evolutionary,lee2023readretro,yuan2024active} select the most promising candidate from multiple predictions offered by the one-step retrosynthesis model at each step of the retrosynthetic planning. This process is repeated until all reactants are commercially available. MCTS~\citep{segler2018planning,hong2021retrosynthetic} enhance search with policy networks for multi-step planning, while DFPN-E~\citep{kishimoto2019depth} combines Depth-First Proof-Number with heuristic techniques for chemical synthesis planning. Neural-based A*-like algorithms in Retro*~\citep{chen2020retro} estimate solution costs to identify promising routes. GRASP~\citep{yu2022grasp} applies RL in guiding the search, and both GNN-Retro~\citep{han2022gnn} and RetroGraph~\citep{xie2022retrograph} use graph neural networks for better cost estimation in A* search. However, we observe that Retro* does not outperform trivial beam search (Retro*-0) in generating feasible synthetic routes as reported in ~\citet{liu2023fusionretro} and we explore a new direction to improve the feasibility of synthetic routes in this work.

\paragraph{Evaluation of Retrosynthetic Planning.}
Some of the existing evaluation~\citep{tu2022retrosynthesis,torren2023models} metrics fall short of verifying whether the starting materials can actually synthesize the given product. \citet{liu2023fusionretro} addresses this issue by extracting synthetic routes from a reaction network and introducing a set-wise exact match evaluation metric. This metric involves comparing the predicted starting materials with the reference starting materials extracted from the reaction database for assessment.

\paragraph{Controllable Generation.} In controllable text generation~\citep{krause2020gedi,yang2021fudge,liu2021dexperts,li2022diffusion}, the language model (LM) remains fixed, while a potential function is used to steer its generation process. These methods utilize various plug-and-play techniques to manage the text generation, ensuring that the produced text meets diverse requirements. Controllable image generation leverages a control network~\citep{zhang2023adding}, to introduce a spatial control signal into large, pre-trained text-to-image diffusion models. This approach enables the generation of images that meet various specific requirements.
\section{Dataset Details}
\label{appendix:dataset}
\begin{table}[htbp]
\centering
\caption{Dataset Statistics.}
\vspace*{\baselineskip}
\label{tab:stat}
\setlength{\tabcolsep}{1mm}
\begin{tabular}{lcccccccccccc}
\toprule
  \diagbox{\textbf{Dataset}}{\textbf{\#Molecules}}{\textbf{Depth}}     & 2  &3 & 4 & 5  & 6   & 7 & 8 & 9 & 10 & 11 &12 & 13   \\
\midrule
Training    & 22,903   & 12,004    & 5,849    &3,268 & 1,432    & 594     & 276     & 107 & 25 & 0 & 0 & 0   \\
Validation    & 2,862   & 1,500    & 731    &408 & 179    & 74     & 34     & 13 & 2 & 0 & 0 & 0   \\
Test    & 2,862   & 1,500    & 731    &408 & 179    & 74     & 34     & 13 & 2 & 32 & 2 & 1   \\
\bottomrule
\end{tabular}
\end{table}
\section{Reproducibility}
\label{appendix:repro}
\subsection{Implementation Details}
\label{appendix:implementation}
We use Pytorch~\citep{paszke2019pytorch} to implement our models. The codes of all baselines are implemented referring to the implementation of FusionRetro~\citep{liu2023fusionretro}. All the experiments of baselines are conducted on a single NVIDIA Tesla A100 with 80GB memory size. The softwares that we use for experiments are Python 3.6.8, CUDA 10.2.89, CUDNN 7.6.5, einops 0.4.1, pytorch 1.9.0, pytorch-scatter 2.0.9, pytorch-sparse 0.6.12, numpy 1.19.2, torchvision 0.10.0, and torchdrug 0.1.3.

\subsection{Hyperparameter Details}
\begin{table}[htbp]
\setlength{\tabcolsep}{0.3mm}
\caption{The hyper-parameters for the forward model.}
\label{tab:hyper_forward}
\vskip 0.15in
    \centering
    \begin{tabular}{l|c}
    \toprule
     max length & 300\\
     embedding size & 64\\
     encoder layers & 6\\
     decoder layers & 6\\
     attention heads & 8\\
     FFN hidden & 1024\\ 
     dropout & 0.1\\
     epochs & 1200\\
     batch size & 256\\
     warmup & 16000\\
     lr factor & 20\\
     scheduling & $lr=\frac{\operatorname{lr\ factor} \times \min \left(1.0, \frac{\operatorname{global\_step}}{\operatorname{warmup}}\right)}{\max\left(\operatorname{global\_step}, \operatorname{warmup}\right)}$\\
    \bottomrule
    \end{tabular}
\end{table}

\begin{table}[htbp]
\setlength{\tabcolsep}{0.3mm}
\caption{The hyper-parameters for the energy-based model.}
\label{tab:hyper_reward}
\vskip 0.15in
    \centering
    \begin{tabular}{l|c}
    \toprule
     max length & 310\\
     embedding size & 64\\
     encoder layers & 6\\
     decoder layers & 6\\
     attention heads & 8\\
     FFN hidden & 1024\\ 
     dropout & 0.1\\
     epochs & 40\\
     batch size & 256\\
     warmup & 10000\\
     lr & 5e-5\\
     scheduling & linear\\
    \bottomrule
    \end{tabular}
\end{table}
\section{A Real-world Example about the Preference of Two Synthetic Routes}
\label{appendix:real example}
Here's a real-world example (generated by ChatGPT) illustrating the trade-off between yield and environmental impact in chemical synthesis:

Example: Synthesis of Acetic Acid.

High-Yield but Less Eco-Friendly Route: Methanol Carbonylation
\begin{itemize}
    \item The industrial production of acetic acid often involves the carbonylation of methanol, catalyzed by iodine and rhodium or cobalt.
    \item Yield: This process, known as the Monsanto or Cativa process, is highly efficient and yields a large quantity of acetic acid.
    \item Environmental Impact: However, it uses significant amounts of metals as catalysts and can produce harmful by-products like iodomethane. The use of these materials and the disposal of by-products can be environmentally damaging.
\end{itemize}

Lower-Yield but More Eco-Friendly Route: Fermentation
\begin{itemize}
    \item An alternative method for acetic acid production is through bacterial fermentation, using species like Acetobacter. This process converts ethanol to acetic acid under aerobic conditions.
    \item Yield: The yield of this method is generally lower compared to the carbonylation of methanol.
    \item Environmental Impact: However, it's considered more environmentally friendly. It uses renewable resources (like ethanol), involves less hazardous chemicals, and generally has a smaller carbon footprint.
\end{itemize}
Decision Making:
Although a company prioritizing production efficiency and cost-effectiveness might choose the carbonylation method for its high yield, despite the greater environmental concerns, we focus on environmental sustainability and prefer the fermentation route despite its lower yield, due to its lesser environmental impact. 

%%%%%%%%%%%%%%%%%%%%%%%%%%%%%%%%%%%%%%%%%%%%%%%%%%%%%%%%%%%%%%%%%%%%%%%%%%%%%%%
%%%%%%%%%%%%%%%%%%%%%%%%%%%%%%%%%%%%%%%%%%%%%%%%%%%%%%%%%%%%%%%%%%%%%%%%%%%%%%%

\end{document}